\documentclass[10pt,twocolumn,letterpaper]{article}

\usepackage{cvpr}              

\usepackage{graphicx}
\usepackage{marvosym}
\usepackage{amsmath}
\usepackage{amssymb}
\usepackage{booktabs}
\usepackage{threeparttable}
\usepackage{bbm}
\usepackage{float}
\usepackage{multirow}
\usepackage{xcolor}
\usepackage{color}
\usepackage[caption=false]{subfig}
\usepackage{colortbl}
\usepackage[accsupp]{axessibility}

\definecolor{mygray}{gray}{0.6}
\definecolor{sgreen}{RGB}{30, 150, 30}
\definecolor{green2}{RGB}{50, 200, 100}
\usepackage[pagebackref=true,breaklinks=true,letterpaper=true,colorlinks,bookmarks=false]{hyperref}

\newcolumntype{x}[1]{>{\centering\arraybackslash}p{#1pt}}

\newlength\savewidth\newcommand\shline{\noalign{\global\savewidth\arrayrulewidth
		\global\arrayrulewidth 1pt}\hline\noalign{\global\arrayrulewidth\savewidth}}
\newcommand{\tablestyle}[2]{\setlength{\tabcolsep}{#1}\renewcommand{\arraystretch}{#2}\centering\footnotesize}
\newcommand\blfootnote[1]{%
  \begingroup
  \renewcommand\thefootnote{}\footnote{#1}%
  \addtocounter{footnote}{-1}%
  \endgroup
}

\usepackage[capitalize]{cleveref}
\crefname{section}{Sec.}{Secs.}
\Crefname{section}{Section}{Sections}
\Crefname{table}{Table}{Tables}
\crefname{table}{Tab.}{Tabs.}

\begin{document}

\title{Progressive Attention on Multi-Level Dense Difference Maps \\ for Generic Event Boundary Detection}

\author{Jiaqi Tang$^1$ \qquad Zhaoyang Liu$^2$ \qquad Chen Qian$^2$ \qquad Wayne Wu$^2$ \qquad Limin Wang\textsuperscript{$1$ \Letter}\\
$^1$State Key Laboratory for Novel Software Technology, Nanjing University, China\\
$^2$SenseTime Research\\
{\tt\small jqtang@smail.nju.edu.cn, zyliumy@gmail.com, \{qianchen, wuwenyan\}@sensetime.com, lmwang@nju.edu.cn}\\
}
\maketitle

\begin{abstract}
Generic event boundary detection (GEBD) is an important yet challenging task in video understanding, which aims at detecting the moments where humans naturally perceive event boundaries. The main challenge of this task is perceiving various temporal variations of diverse event boundaries. To this end, this paper presents an effective and end-to-end learnable framework (DDM-Net). To tackle the diversity and complicated semantics of event boundaries, we make three notable improvements. First, we construct a feature bank to store multi-level features of space and time, prepared for difference calculation at multiple scales. Second, to alleviate inadequate temporal modeling of previous methods, we present dense difference maps (DDM) to comprehensively characterize the motion pattern. Finally, we exploit progressive attention on multi-level DDM to jointly aggregate appearance and motion clues. As a result, DDM-Net respectively achieves a significant boost of 14\% and 8\% on Kinetics-GEBD and TAPOS benchmark, and outperforms the top-1 winner solution of LOVEU Challenge@CVPR 2021 without bells and whistles. The state-of-the-art result demonstrates the effectiveness of richer motion representation and more sophisticated aggregation, in handling the diversity of GEBD. The code is made available at \url{https://github.com/MCG-NJU/DDM}.
\end{abstract}
\blfootnote{\Letter: Corresponding author.}

\section{Introduction}
With the explosive growth of online videos, video understanding has drawn tremendous attention from both academia and industry. Cognitive science~\cite{tversky2013event} suggests that humans naturally divide a video into meaningful units by perceiving event boundaries. To this end, a task termed as \textbf{Generic Event Boundary Detection}~\cite{Shou_2021_ICCV} (GEBD) is recently proposed to localize the generic event boundaries in videos, which is expected to facilitate the development of video understanding. 

\begin{figure}[t]
\begin{center}
\includegraphics[scale=0.23]{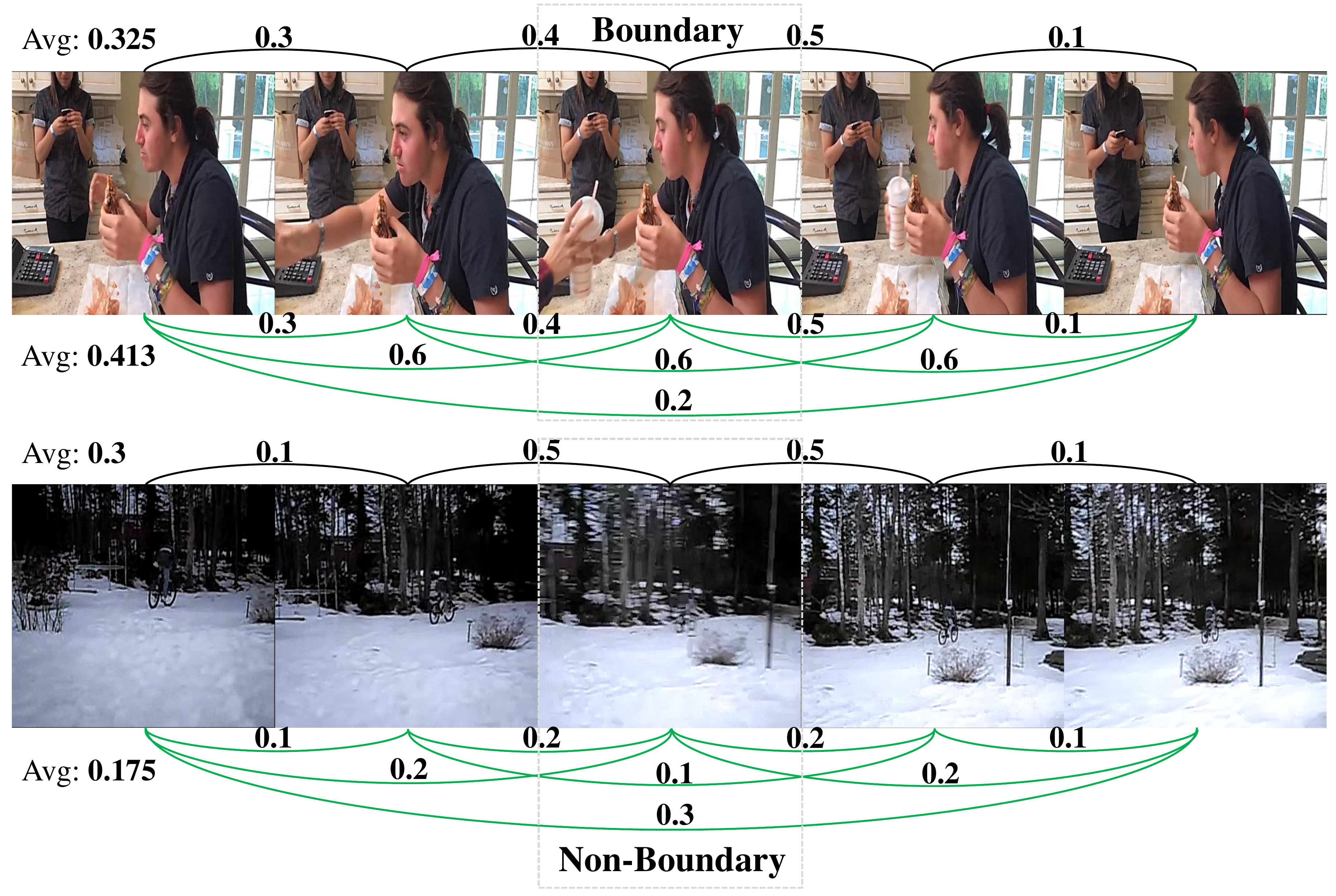}
\vspace{-2mm}
\caption{\textbf{Comparisons of sparse motion representation} (black lines, optical flow) \textbf{and dense motion representation} (green lines, some are omitted for clarity, dense feature differences). Numbers on lines indicate the magnitude of motion between two frames. Dense motion representation provides more holistic temporal cues to better distinguish boundaries and non-boundaries.}
\label{fig:intro}
\end{center}
\vspace{-4mm}
\end{figure}

Generic event boundaries in GEBD task are taxonomy-free and related to a broad range of temporal changes, including changes of action, subject and environment. 
The primary challenge in GEBD task is to model diverse patterns of generic event boundaries:  \emph{a) Spatial diversity} is dominantly characterized by the change of appearance, which normally comprises low-level changes (\eg, change in color or brightness) and high-level changes (\eg, the dominant subject appears or disappears). \emph{b) Temporal diversity} is mainly relevant to actions, such as change of action (\eg, walk to run) or change of object of interaction. Notably, different actions usually exhibit inconsistent speed and duration, which further increases the temporal diversities of event boundaries. 
As a result, the spatio-temporal diversities lead to overly complicated variations in videos, which impedes the accurate detection of event boundaries.

Since GEBD task is highly correlated with changes in temporal dimension, motion information is the key to perceiving temporal variations and detecting event boundaries. Previous methods wildly use optical flow~\cite{DBLP:conf/eccv/WangXW0LTG16, DBLP:conf/eccv/LinZSWY18, DBLP:conf/iccv/LinLLDW19} as alternative motion representation to learn temporal clues in videos. However, they model the semantics in a single feature level and focus on local motion cues between two consecutive frames (Figure~\ref{fig:intro}), which is insufficient to perceive diverse event boundaries. In addition, previous two-stream methods~\cite{DBLP:conf/nips/SimonyanZ14, DBLP:conf/eccv/WangXW0LTG16} commonly resort to simple fusion schemes, short of interaction across appearance and motion modalities. Hence, they are less effective for learning complex semantics of diverse event boundaries. 

\begin{figure*}[t]
\vspace{-2mm}
\begin{center}
\includegraphics[scale=0.66]{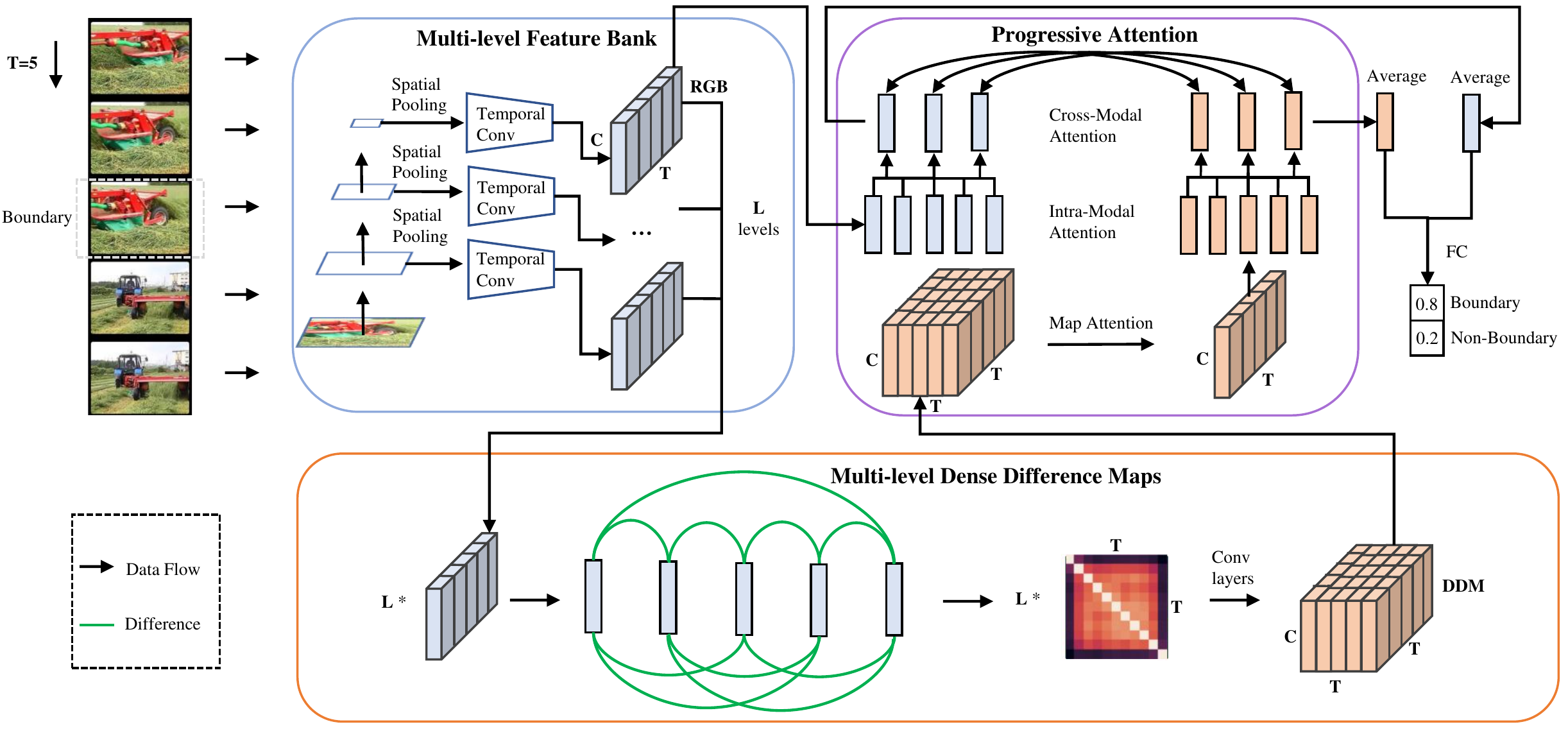}
\end{center}
\vspace{-6mm}
   \caption{{\bf Overview of DDM-Net.} Our DDM-Net streamlines the process of generic event boundary detection by viewing it as a binary classification problem of sliding video clips. Specifically, our method classifies the current frame with a clip centered on it and repeats the same process on other frames. The network is mainly composed of three stages: multi-level feature bank construction, dense difference map calculation, and progressive attention. DDM-Net exploits richer motion information and more sophisticated aggregation to achieve accurate detection for generic event boundaries. (L: number of levels of features, T: number of frames, C: number of channels.)}
\label{fig_overview}
\vspace{-4mm}
\end{figure*}

To address the above issues, we present a method (DDM-Net) that progressively aggregates dense motion information along with appearance cues to perceive event boundaries, as illustrated in Figure~\ref{fig_overview}. We make three notable improvements, including Multi-Level Feature Bank, Dense Difference Map and Progressive Attention. \emph{First}, we build a \emph{Multi-Level Feature Bank} where the features are collected in different spatial and temporal scales respectively, which empowers the subsequent modules to thoroughly perceive different levels of changes in videos.

\emph{Second}, based on aforementioned feature bank, we propose a \emph{Dense Difference Map} (DDM) to model rich temporal contexts. Technically, we calculate pairwise feature differences between every two frames in a clip of length $T$, and obtain a $T\times T$ dense difference map. The main advantage of DDM is to exploit the difference of each feature pair and provide holistic motion information. As shown in Figure~\ref{fig:intro}, our proposed DDM is able to provide more holistic and salient temporal clues than optical flow, which is calculated between two consecutive frames. Furthermore, instead of directly being operated on raw frames, our DDM is built on the features collected from different layers of backbone network, and thus ought to be more robust to temporal noise (\eg, camera blur in the second row of Figure~\ref{fig:intro}).

\emph{Third}, as event boundaries show their spatio-temporal diversities and complexities, we argue that simple linear fusion in two-stream methods is insufficient to aggregate the appearance and motion clues. We thus exploit \emph{Progressive Attention} to mine important clues hidden in RGB features and DDM. In order to align the shape of DDM to RGB features, we design map-squeezed attention to squeeze DDM. Then, in intra-modal attention, key features of two modalities are respectively enhanced through two sets of learnable queries, prepared for cross-modal attention. Cross-modal attention is leveraged to perform feature interaction across modalities, enabling appearance and motion features to query and guide each other. As a result, DDM-Net can more effectively aggregate spatio-temporal clues and improve the discrimination of event boundaries.

Our DDM-Net exploits multi-level dense differences to perceive diverse temporal variations, and leverages progressive attention to effectively aggregate appearance and motion clues. To prove the effectiveness of DDM-Net, we perform extensive experiments on two datasets: Kinetics-GEBD~\cite{Shou_2021_ICCV} and TAPOS~\cite{DBLP:conf/cvpr/ShaoZDL20}. Evaluation results demonstrate that our DDM-Net outperforms the existing state-of-the-art methods by a large margin on all evaluation metrics. Particularly, DDM-Net obtains a superior 76.4\% F1@0.05 on Kinetics-GEBD, with a significant boost of 14 percent. On TAPOS, we improve F1 score@0.05 from 52.2\% to 60.4\%. 
In addition, our DDM-Net is superior to winners of LOVEU Challenge@CVPR 2021~\cite{Shou_2021_ICCV} on the testing set of Kinetics-GEBD, demonstrating the effectiveness of our method. In summary, our main contributions are as follows:
\begin{itemize}
    \item We propose dense difference maps equipped with multi-level feature bank to leverage richer temporal clues for detection of diverse event boundaries.
    
    \item Instead of simple feature fusion methods, progressive attention is employed to aggregate appearance and motion clues from RGB features and DDM, enabling DDM-Net to generate more discriminative representations and learn more complicated semantics. 
    
    \item Extensive experiments and studies demonstrate that our DDM-Net achieves the state-of-the-art performance on Kinetics-GEBD and TAPOS benchmark, under the setting of the same backbone.
    
\end{itemize}

\section{Related Work}
\noindent\textbf{Temporal Detection Tasks in Video Understanding.}
Temporal action detection task aims to detect action instances in untrimmed videos, namely predict starting point, ending point and category of each action. One-stage and two-stage methods are two mainstream solutions. Different from direct one-stage methods~\cite{DBLP:conf/bmvc/BuchEGFN17, DBLP:conf/cvpr/XuZRTG20, DBLP:conf/cvpr/Lin0LWTWLHF21}, two-stage methods~\cite{DBLP:conf/eccv/LinZSWY18, DBLP:conf/iccv/LinLLDW19, DBLP:conf/iccv/ZengHGTRZH19, DBLP:conf/cvpr/QingSGW0W0YGS21, DBLP:conf/cvpr/LiuHBDBT21, Tan_2021_ICCV} decompose the task into class-agnostic proposal generation and action classification. Temporal action parsing~\cite{DBLP:conf/cvpr/ShaoZDL20} is recently proposed, of which the target is dividing actions into segments of sub-actions. Video anomaly detection~\cite{DBLP:conf/cvpr/LiuLLG18, DBLP:conf/cvpr/SultaniCS18, DBLP:conf/iccv/GongLLSMVH19, DBLP:conf/eccv/LuYR020} is aimed at recognizing frames where abnormal events happen, wildly applied in video surveillance. As for shot boundary detection~\cite{DBLP:conf/caip/BaraldiGC15, DBLP:conf/cbmi/Gygli18, DBLP:conf/accv/TangFKCZ18}, it is a classical task for significant shot change detection. Different from them, GEBD~\cite{Shou_2021_ICCV} is a generic detection task, where generic event boundaries include all of the above. To address the diversity and complicated semantics of generic event boundaries, our method improves the boundary discrimination via progressively attending to multi-level dense difference maps.

\noindent\textbf{Motion Representation.}
Previous methods of current video understanding tasks (\eg, action recognition, temporal action detection, \etc) wildly used
optical flow~\cite{DBLP:conf/nips/SimonyanZ14, DBLP:conf/eccv/WangXW0LTG16, DBLP:conf/cvpr/CarreiraZ17, DBLP:conf/iccv/LinLLDW19}, RGB differences~\cite{DBLP:journals/pami/JunejoDLP11, DBLP:conf/eccv/WangXW0LTG16, DBLP:conf/cvpr/ZhaoXL18, DBLP:conf/iccv/LuoY19, Zhi_2021_ICCV} and feature differences~\cite{DBLP:conf/aaai/LiuLWWTWLHL20, DBLP:conf/iccv/JiangWGWY19, DBLP:conf/cvpr/LiJSZKW20, DBLP:conf/cvpr/DwibediATSZ20, Liu_2021_ICCV, DBLP:conf/cvpr/0002TJW21} as the motion representation to learn temporal clues in videos. However, they focus on local motion cues between two consecutive frames and do not explicitly employ multi-level features for complex semantics learning. Compared with single-level sparse motion representations, our proposed multi-level DDM is a dense motion representation built upon the multi-level feature bank, better to perceive diverse temporal variations in multiple levels.

\noindent\textbf{Multi-Modal Feature Aggregation and Fusion.}
Multi-modal feature aggregation and fusion are wildly applied for holistic semantics learning in many tasks, such as image-text~\cite{DBLP:conf/icml/XuBKCCSZB15, DBLP:conf/cvpr/YangHGDS16, DBLP:conf/cvpr/00010BT0GZ18, DBLP:conf/nips/LuBPL19}, video-text~\cite{DBLP:conf/emnlp/LeiYBB18, DBLP:conf/iccv/GaoSYN17, DBLP:conf/iccv/HendricksWSSDR17} and audio-video tasks~\cite{DBLP:conf/eccv/GaoFG18, DBLP:conf/eccv/TianSLDX18}. In video understanding, previous two-stream methods~\cite{DBLP:conf/nips/SimonyanZ14, DBLP:conf/cvpr/FeichtenhoferPZ16, DBLP:conf/eccv/WangXW0LTG16} train two separate networks and resort to a simple fusion of two video modalities, namely appearance and motion, via linear fusion or feature concatenation. However, they are less effective due to a lack of interaction across two modalities, falling short of dependencies between appearance and motion features~\cite{chen2021endtoend}. To this end, our method progressively attends to multi-level DDM, taking advantage of correlations between two modalities to enrich the semantic information and improve the discrimination.

\section{Method}
\subsection{Overview}
Generic Event Boundary Detection~\cite{Shou_2021_ICCV} (GEBD) aims to detect the taxonomy-free event boundaries, \eg, change of action, change of subject, shot change, etc. As the temporal boundaries in video usually exhibit the dominant characteristics of ambiguity and diversity, it indeed is a challenging vision task remaining to be studied. To this end, we propose a novel spatio-temporal modeling scheme, which constructs and attends to multi-level dense difference maps to address aforementioned issues.

Given a video $V =\{I_t\}_{t=1}^{E}$, where $I_t$ is the $t$-$th$ frame and $E$ is the number of frames in video, we sample a clip $U=\{I_{t-w}, ..., I_t, ..., I_{t+w}\}$ of $T$ ($T=2\times w+1$) frames from video $V$ to infer whether $I_t$ is a boundary frame. 
As illustrated in Figure~\ref{fig_overview}, DDM-Net mainly refers to three parts: a multi-level spatio-temporal feature bank, multi-level dense difference maps and  cross-modal aggregation between \textbf{RGB features} $A$ and \textbf{Dense Difference Map} (DDM) $M$ via progressive attention. \emph{Firstly}, the sampled clip $U$ is fed into a backbone network and a serial of temporal convolutions to yield \emph{Multi-Level} spatio-temporal features $\mathbf{F} = \{f_{ij}\}_{i\in[1,m], j\in[1,n]}$, where $i$ and $j$ respectively denotes the spatial level ($m$ levels in total) and the temporal level ($n$ levels in total) of the feature.
\emph{Secondly}, a \emph{Dense Difference Map} $M\in \mathbb{R}^{C\times T\times T}$ is constructed with $\mathbf{F}$ by measuring the discrepancy among frames, which is expected to provide more discriminative information to aid the model in perceiving temporal variations. 
\emph{Thirdly}, in our \emph{Progressive Attention} module, intra-modal attention module exploits a set of learnable queries to enhance key intra-modal representations, and co-attention transformers are leveraged to perform cross-modal attention. It is worth noting that, to align with RGB features $A$, DDM $M\in \mathbb{R}^{C\times T\times T}$ is firstly squeezed to a sequence $D \in \mathbb{R}^{C\times T}$ via map-squeezed attention.
\emph{Finally}, $A$ and $D$ are respectively fed into separate fully-connected ($fc$) layers. With a linear fusion after $fc$ layers, the model outputs the final boundary probability of the center frame $I_t$. 

In contrast to previous two-stream methods~\cite{DBLP:conf/nips/SimonyanZ14, DBLP:conf/cvpr/FeichtenhoferPZ16, DBLP:conf/eccv/WangXW0LTG16} that use optical flow as motion representation to learn temporal clues in videos, we meticulously construct dense difference maps, which enable the model to perceive generic event boundaries along with RGB features. Since DDM is calculated on-the-fly, our method is more efficient than previous two-stream methods that train two separate networks. In the following sections, we will introduce the technical details of each module.
\label{section:overview}

\subsection{Multi-Level Feature Bank}
To model diverse motion patterns of generic event boundaries, we exploit a feature bank to store multi-level features of input video clips, based on which the dense difference map is calculated to yield rich temporal clues.

\noindent\textbf{Temporal View of Multi-Level Feature Bank.} Before building the feature bank, an issue that needs to be figured out is whether we take a clip or the whole video as inputs to detect event boundaries. As videos normally are composed of multiple non-overlapping and relatively independent snippets that belong to different events, we argue that whether the current frame is an event boundary is mostly related to its adjacent snippets. Snippets far away from the current frame contribute little to infer whether it is an event boundary. 
Therefore, we opt to build our model based on a clip around the current frame, instead of the whole video. Notably, experiments in Table~\ref{table:view} have also examined the rationality of our point.
Specifically, along with the current frame, we sample $w$ frames before and after the current frame, namely $T$ ($T=2 \times w + 1$) frames as an input clip. Then, the input clip is fed into backbone network to construct the multi-level feature bank. 

\noindent\textbf{Construction of Multi-Level Feature Bank.} Since event boundaries in GEBD task are generic and taxonomy-free, patterns of different event boundaries vary considerably in space and time. From a perspective of space, appearance changes include low-level changes and high-level changes. Low-level changes mainly refer to change in environment (\ie, change in color and brightness), while high-level changes are related to complex semantics (\eg, the dominant subject appears or disappears). From a temporal perspective, the duration of action changes is usually inconsistent. For instance, `a runner suddenly changes direction' can happen very fast, while `an old man slowly stands up' usually takes several frames. 

To detect event boundaries with diverse motion patterns, our method models temporal variations upon multi-level spatio-temporal features. Specifically, we perform average spatial pooling on $m$ layers of ResNet features (\eg, layer3 and layer4), and get $m$ feature sequences of different semantic levels. It is notable that the feature sequence of high-level layer4 is also denoted as RGB features $A$, which are later fused with DDM features $D$, as shown in Figure~\ref{fig_overview}. Then, for each feature sequence, we exploit temporal convolutions to get $n$ feature sequences with different temporal receptive fields. Consequently, there are $m\times n=L$ levels of features in total, prepared for multi-level dense difference calculation. In Table~\ref{table:multilevel}, we observe that both multi-level features from spatial and temporal domain provide crucial clues to detect diverse event boundaries.
\label{section:multilevel}

\subsection{Dense Difference Maps}
Motion representation is crucial in GEBD task. As for boundaries like changes of action, there is little change in appearance (\eg, a man waves gently towards the camera, or walk to run). To detect such boundaries, motion information plays a principal role in perceiving temporal variations. Previous methods commonly exploit sequential optical flow or RGB differences to approximate motion information. However, they can only reflect local motion cues between two consecutive frames and fail to take advantage of rich temporal contexts. Considering the variety of boundaries and complicated scenarios in GEBD task, it is insufficient to use local and sparse motion representation. 

To alleviate inadequate temporal context modeling of sparse motion representation, we propose dense difference maps based on aforementioned multi-level feature bank. Given a feature sequence of $T$ frames, we calculate the feature difference of each frame pair and construct a $T\times T$ map. Compared with the sparse motion sequence of length $T-1$, $T\times T$ pairs of feature differences provide denser temporal cues (Figure~\ref{fig:intro}). Since DDM contains richer motion information, it characterizes the motion pattern around the current frame more holisticly, enabling our method to better perceive temporal variations and distinguish boundaries and non-boundaries. Moreover, DDM is constructed with aforementioned multi-level feature bank, where features are collected from different layers of backbone network and consist of 
multi-level semantics. Hence, it is more robust to temporal noise than optical flow and RGB differences, which are directly calculated on raw frames.  

In practice, Euclidean distance is employed across all the channels to measure the feature difference between two frames $I_i$ and $I_j$, 
\vspace{-2mm}
\begin{equation}
    FD(i, j) = \sqrt{\sum_{c=1}^{C}(A_{i}^c - A_{j}^c)^2},
\label{Class_loss}
\end{equation}
where $A_{i}$ and $A_j$ are appearance features of $I_i$ and $I_j$, $C$ is the total number of channels. Then, we exploit stacked convolution layers to transform difference matrices$\in \mathbb{R}^{L\times T\times T}$ into $M\in \mathbb{R}^{C\times T\times T}$. In Table~\ref{table:operator}, DDM-Net also achieves close performance with other distance metrics(\eg, Manhattan distance), which demonstrates the performance of our method is robust to the choice of difference operators.

\subsection{Progressive Attention}
Previous two-stream networks usually leverage simple aggregation and fusion manners, such as linear fusion or feature concatenation of temporal averaging results. However, they lack interaction between modalities and thus cannot take full advantage of our proposed DDM, which is proved in Table~\ref{table:progressiveattn}. Hence, to better aggregate appearance and motion clues, we employ progressive attention on our proposed multi-level DDM, including map-squeezed attention, intra-modal attention and cross-modal attention.

\vspace{2mm}
\noindent\textbf{Map-Squeezed Attention.}
To align $M\in \mathbb{R}^{C\times T\times T}$ with RGB features $A\in \mathbb{R}^{C\times T}$, we transform it into a  feature sequence of length $T$ via frame-wise map-squeezed attention. In DDM, feature sequence of the i-th row ($M_i\in \mathbb{R}^{C\times T}$) is the difference between the i-th frame $I_i$ and other frames of the current clip. Hence, it is intuitive to aggregate elements of the $M_i$ to get a clip-level motion measurement of the $I_i$. Due to the diversity of temporal dependencies, it is common that differences with several specific frames are more important than others. Therefore, we propose a frame-wise attention mechanism to squeeze $M$, calculating the weights of all elements in $M_i$ based on feature $A_{i}$ of $I_i$. Concretely, we exploit $A_{i}$ to attend all elements of $M_i$ and generate weights $\gamma_i$, adaptively aggregating all differences into a motion measurement $D_i$, formulated as:
\begin{equation}
\begin{aligned}
& {{\mu}_{ij}=\mathbf{W}_{\mu}^{\intercal} \left(\mathbf{W}_{A}^{\intercal} A_{i}+\mathbf{W}_{M}^{\intercal} M_{ij}\right)}, \\
& \label{mapattn}
{{\gamma}_{i j}=\frac{\exp \left({\mu}_{i j}\right)}{\sum\nolimits_{t=1}^{T} \exp \left({\mu}_{i t}\right)}}, \\ 
& D_i = \sum\nolimits_{j=1}^{T} {\gamma}_{i j} M_{ij},\\
\end{aligned}
\end{equation}
where $\mathbf{W}_{A}^{\intercal}$, $\mathbf{W}_{M}^{\intercal}$ and $\mathbf{W}_{\mu}^{\intercal}$ are projection matrices.

\vspace{2mm}
\noindent\textbf{Intra-Modal Attention.}
As mentioned in Section~\ref{section:overview} and~\ref{section:multilevel}, our method predicts the boundary confidence of the current frame $I_t$ based on a clip $U$ centered on it. In the clip, features of different timestamps should not be equally important. For example, the center frame of the clip is more important than edge frames of the clip in most cases. To adaptively aggregate and enhance key representations of RGB features $A$ and DDM features $D$, we employ two sets of $\omega$ \emph{learnable} queries $\mathbf{q}$, which are formed by adding content queries $\mathbf{c}_q$ (initialized with standard normal distribution) and learnable positional embeddings of queries $\mathbf{p}_q$. Specifically, We exploit two separate transformer decoders to respectively aggregate and enhance key intra-modal features of $A$ and $D$, 
\begin{equation}
\begin{aligned}
    & \mathbf{q} = \mathbf{c}_q + \mathbf{p}_q,\\
    & \mathbf{k} = \mathbf{c}_k + \mathbf{p}_k = H + \mathbf{p}_k,
    & \mathbf{v} = \mathbf{c}_v = H,\\
\end{aligned}
\end{equation}
where $\mathbf{c}_k$ and $\mathbf{c}_v$ are features $H$ of the modality ($A$ or $D$), $\mathbf{p}_k$ is sine positional embedding.  In cross-attention layers, queries globally attend and aggregate features of high activation into each query. Self-attention layers model the dependencies between queries and enhance corresponding query embeddings. Through intra-modal representation learning, two sets of queries $\mathbf{q}$ independently aggregate and enhance key features of two modalities, and become refined queries $\mathbf{q}^{'}$. In Table~\ref{table:progressiveattn}, we observe that cross-modal attention can achieve better performance upon refined key features $\mathbf{q}^{'}$, compared with the unrefined features $H$.

\vspace{2mm}
\noindent\textbf{Cross-Modal Attention.}
Due to the diversity and complex semantics of generic event boundaries, it is difficult to distinguish them with only appearance or motion features. A fusion of them can alleviate this issue, but previous fusion methods (\eg, feature concatenation) fail to jointly learn features across modalities and make full use of feature complementarity. Thus, in order to leverage the dependencies between two modalities, we perform cross-modal feature aggregation. Concretely, we take the \emph{feature pair} of $\omega$ refined queries $\mathbf{q}^{'}$ as the input of two independent co-attention transformers.
One co-attention transformer takes refined RGB features $\mathbf{q}_{A}^{'}$ as queries, and refined DDM features $\mathbf{q}_{D}^{'}$ as keys and values, 
\begin{equation}
\begin{aligned}
    & \mathbf{q} = \mathbf{c}_q = \mathbf{q}_{A}^{'},\\
    & \mathbf{k} = \mathbf{c}_k = \mathbf{q}_{D}^{'},
    & \mathbf{v} = \mathbf{c}_v = \mathbf{q}_{D}^{'}.\\
\end{aligned}
\end{equation} 
That is to say, $\mathbf{q}_{A}^{'}$ guide and enhance $\mathbf{q}_{D}^{'}$ via cross-attention layers. Inputs of the other co-attention transformer are symmetric to the first one, namely $\mathbf{q}_{D}^{'}$ as queries, $\mathbf{q}_{A}^{'}$ as keys and values. Through cross-attention layers, cross-modal attention module outputs RGB-conditioned DDM features $\mathbf{q}_{D}^{''}$ and DDM-modulated RGB features $\mathbf{q}_{A}^{''}$. As a consequence, DDM-Net aggregates appearance and motion cues with cross-modal guidance, effectively improving the discrimination of event boundaries.

\begin{table*}[t]
\centering
\begin{center}
\scalebox{0.9}{
\begin{tabular}{ccccccccccccc}                                                                                            &                \\ \toprule[1pt]
                                               \multicolumn{2}{c|}{Rel.Dis. threshold} & 0.05           & 0.1            & 0.15           & 0.2            & 0.25           & 0.3            & 0.35           & 0.4            & 0.45           & \multicolumn{1}{c|}{0.5}            & avg            \\ \hline
\multicolumn{1}{c|}{\multirow{3}{*}{Unsuper.}} & \multicolumn{1}{c|}{SceneDetect~\cite{castellano2014PySceneDetect}}    & 0.275 & 0.300          & 0.312          & 0.319          & 0.324          & 0.327          & 0.330          & 0.332          & 0.334          & \multicolumn{1}{c|}{0.335}          & 0.318          \\
\multicolumn{1}{c|}{}                          & \multicolumn{1}{c|}{PA - Random~\cite{Shou_2021_ICCV}}    & 0.336          & 0.435          & 0.484         & 0.512          & 0.529          & 0.541          & 0.548          & 0.554          & 0.558          & \multicolumn{1}{c|}{0.561}          & 0.506          \\
\multicolumn{1}{c|}{}                          & \multicolumn{1}{c|}{PA~\cite{Shou_2021_ICCV}}      & 0.396          & 0.488 & 0.520 & 0.534 & 0.544 & 0.550 & 0.555 & 0.558 & 0.561 & \multicolumn{1}{c|}{0.564} & 0.527
\\ \hline

\multicolumn{1}{c|}{\multirow{6}{*}{Super.}}   & \multicolumn{1}{c|}{BMN~\cite{DBLP:conf/iccv/LinLLDW19}}            & 0.186        & 0.204 & 0.213 & 0.220 & 0.226 & 0.230 & 0.233 & 0.237 & 0.239 & \multicolumn{1}{c|}{0.241}          &  0.223   \\
\multicolumn{1}{c|}{}   & \multicolumn{1}{c|}{BMN-StartEnd~\cite{DBLP:conf/iccv/LinLLDW19}}            & 0.491 & 0.589 & 0.627 & 0.648 & 0.660 & 0.668 & 0.674 & 0.678 & 0.681 & \multicolumn{1}{c|}{0.683}          &  0.640  \\
\multicolumn{1}{c|}{}                          & \multicolumn{1}{c|}{TCN-TAPOS~\cite{DBLP:conf/eccv/LeaRVH16}}            & 0.464  & 0.560  & 0.602  & 0.628  & 0.645  & 0.659  & 0.669  & 0.676  & 0.682  & \multicolumn{1}{c|}{0.687}          &   0.627
\\ 
\multicolumn{1}{c|}{}                          & \multicolumn{1}{c|}{TCN~\cite{DBLP:conf/eccv/LeaRVH16}}            & 0.588 & 0.657 & 0.679 & 0.691 & 0.698 & 0.703 & 0.706 & 0.708 & 0.710 & \multicolumn{1}{c|}{0.712}          &   0.685
\\
\multicolumn{1}{c|}{}                          & \multicolumn{1}{c|}{PC~\cite{Shou_2021_ICCV}}            & 0.625  & 0.758  & 0.804  & 0.829  & 0.844  & 0.853  & 0.859  & 0.864  & 0.867  & \multicolumn{1}{c|}{0.870}          &   0.817
\\
\cline{2-13}
\multicolumn{1}{c|}{}                          & \multicolumn{1}{c|}{DDM-Net}            & \textbf{0.764}  & \textbf{0.843}  & \textbf{0.866}  & \textbf{0.880}  & \textbf{0.887}  & \textbf{0.892}  & \textbf{0.895}  & \textbf{0.898}  & \textbf{0.900}  & \multicolumn{1}{c|}{\textbf{0.902}}          &   \textbf{0.873}
\\ \bottomrule[1pt]
\end{tabular}
}
\end{center}
\vspace{-5mm}
\caption{Comparison with previous methods on the validation set of Kinetics-GEBD, measured by F1 score at different Rel.Dis. thresholds.}
\vspace{-2mm}
\label{table:GEBD}
\end{table*}

\begin{table*}[t]
\centering
\scalebox{0.9}{
\begin{tabular}{ccccccccccccc}                                                                                            &                \\ \toprule[1pt]
                                               \multicolumn{2}{c|}{Rel.Dis. threshold} & 0.05           & 0.1            & 0.15           & 0.2            & 0.25           & 0.3            & 0.35           & 0.4            & 0.45           & \multicolumn{1}{c|}{0.5}            & avg            \\ \hline
\multicolumn{1}{c|}{\multirow{3}{*}{Unsuper.}} & \multicolumn{1}{c|}{SceneDetect~\cite{castellano2014PySceneDetect}}    & 0.035          & 0.045          & 0.047          & 0.051          & 0.053          & 0.054          & 0.055          & 0.056          & 0.057          & \multicolumn{1}{c|}{0.058}          & 0.051          \\
\multicolumn{1}{c|}{}                          & \multicolumn{1}{c|}{PA - Random~\cite{Shou_2021_ICCV}}   & 0.158          & 0.233          & 0.273          & 0.310          & 0.331          & 0.347          & 0.357          & 0.369          & 0.376          & \multicolumn{1}{c|}{0.384}          &    0.314        \\
\multicolumn{1}{c|}{}                          &
\multicolumn{1}{c|}{PA~\cite{Shou_2021_ICCV}}      & 0.360 & 0.459 & 0.507 & 0.543 & 0.567 & 0.579 & 0.592 & 0.601 & 0.609 & \multicolumn{1}{c|}{0.615} & 0.543 \\
\hline

\multicolumn{1}{c|}{\multirow{6}{*}{Super.}}   & \multicolumn{1}{c|}{ISBA~\cite{DBLP:conf/cvpr/DingX18}}           & 0.106          & 0.170          & 0.227          & 0.265          & 0.298          & 0.326          & 0.348          & 0.369          & 0.382          & \multicolumn{1}{c|}{0.396}          & 0.302          \\
\multicolumn{1}{c|}{}                  & \multicolumn{1}{c|}{TCN~\cite{DBLP:conf/eccv/LeaRVH16}}            & 0.237          & 0.312          & 0.331          & 0.339          & 0.342          & 0.344          & 0.347          & 0.348          & 0.348          & \multicolumn{1}{c|}{0.348}          & 0.330                  \\
\multicolumn{1}{c|}{}                          & \multicolumn{1}{c|}{CTM~\cite{DBLP:conf/eccv/HuangFN16}}            & 0.244          & 0.312          & 0.336          & 0.351          & 0.361          & 0.369          & 0.374          & 0.381          & 0.383          & \multicolumn{1}{c|}{0.385}          & 0.350          \\
\multicolumn{1}{c|}{}                          & \multicolumn{1}{c|}{TransParser~\cite{DBLP:conf/cvpr/ShaoZDL20}}    & 0.289          & 0.381          & 0.435          & 0.475          & 0.500       & 0.514          & 0.527          & 0.534          & 0.540          & \multicolumn{1}{c|}{0.545}          & 0.474          \\
\multicolumn{1}{c|}{}                          & \multicolumn{1}{c|}{PC~\cite{Shou_2021_ICCV}}    & 0.522          & 0.595          & 0.628          & 0.646          & 0.659       & 0.665          & 0.671          & 0.676          & 0.679          & \multicolumn{1}{c|}{0.683}          & 0.642
\\
\cline{2-13}
\multicolumn{1}{c|}{}                          & \multicolumn{1}{c|}{DDM-Net}            & \textbf{0.604}  & \textbf{0.681}  & \textbf{0.715}  & \textbf{0.735}  & \textbf{0.747}  & \textbf{0.753}  & \textbf{0.757}  & \textbf{0.760}  & \textbf{0.763}  & \multicolumn{1}{c|}{\textbf{0.767}}          &   \textbf{0.728}
\\ \bottomrule[1pt]
\end{tabular}
}
\vspace{-2mm}
\caption{Comparison with previous GEBD methods on the validation set of TAPOS, measured by F1 score at different Rel.Dis. thresholds.}
\vspace{-2mm}
\label{table:TAPOS}
\end{table*}

\begin{table}
\begin{center}
\scalebox{0.85}{
\begin{tabular}{ccccc}
\toprule[1pt]
Method & rank1~\cite{kang2021winning} & rank2~\cite{hong2021generic} & rank3~\cite{sun2021generic} & DDM-Net \\ 
\hline
F1 score & 0.8354 & 0.8330 & 0.8309 & \textbf{0.8368} \\ 
\bottomrule[1pt]
\end{tabular}
}
\vspace{-2mm}
\caption{Comparison with winner solutions of LOVEU challenge on the testing set of Kinetics-GEBD, measured by F1 score@0.05.}
\label{table:loveu}
\vspace{-6mm}
\end{center}
\end{table}

\begin{table*}[t]\centering
		\captionsetup[subfloat]{captionskip=1pt}
		\captionsetup[subffloat]{justification=centering}
		\subfloat[\textbf{Study on Different Representations.} We compare the performance of single modality and two modalities. When combined with RGB features, only DDM is calculated online and can be trained on-the-fly.
		\label{table:representation}]{
		\scalebox{0.9}{
			\tablestyle{4pt}{1.05}
			\begin{tabular}{x{140}|x{25}x{25}x{25}x{25}}
         Representation & 0.05 & 0.25 & 0.5 & Average \\ 
\shline
RGB & 0.6793 & 0.8589 &	0.8772 & 0.8375 \\
Optical flow & 0.6625 &	0.8045 & 0.8206 & 0.7877 \\
RGB differences & 0.7272 & 0.8591 &	0.8753 & 0.8440 \\
DDM & 0.7512 &	0.8738 & 0.8861 & 0.8591 \\
\shline
RGB~+~Optical flow~(two-stream) & 0.6881 & 0.8682 & 0.8844 &	0.8465 \\
RGB~+~RGB differences~(two-stream) & 0.7307 & 0.8702 & 0.8834 & 0.8536 \\
RGB~+~DDM~(on-the-fly) & \textbf{0.7643} & \textbf{0.8870} &	\textbf{0.9016} & \textbf{0.8726} \\
         \multicolumn{5}{c}{}\\
        \end{tabular}}} \hspace{10mm}
		\subfloat[\textbf{Study on Temporal Views.} We compare the performance of input clips with different temporal views,  and further boost the performance with a more-frames setting.
		\label{table:view}]{
		\scalebox{0.9}{
			\tablestyle{4pt}{1.05}
			 \begin{tabular}{c|c|x{25}x{25}x{25}x{25}}
         $w$ & $s$ & 0.05 & 0.25 & 0.5 & Average\\
         \shline
        5 & 3 & 0.7551 & 0.8815 & 0.8970 & 0.8666 \\
5 & 6 & \textbf{0.7643} & \textbf{0.8870} &	\textbf{0.9016} & \textbf{0.8726} \\
5 & 12 & 0.7521 & 0.8788 & 0.8924 & 0.8636 \\
5 & 30 & 0.6914 & 0.8563 & 0.8717 & 0.8371 \\
\shline
15 & 2 & \textbf{0.7703} & \textbf{0.8890} & \textbf{0.9042} & \textbf{0.8754} \\
         \multicolumn{6}{c}{}\\
        \end{tabular}}}
        
		\subfloat[\textbf{Study on Multi-Level Feature Bank.} 
	`None' refers to only using the feature collected from the last layer of backbone network (layer4 of ResNet50). 
		\label{table:multilevel} ]{
		\scalebox{0.9}{
			\tablestyle{2pt}{1.05}
			\begin{tabular}{x{61}|x{25}x{25}x{25}x{25}}
         Multi-level & 0.05 & 0.25 & 0.5 & Average \\
         \shline
         None & 0.7353 & 0.8617 & 0.8726 & 0.8463 \\
Spatial  & 0.7487 & 0.8694 & 0.8820 & 0.8552 \\
Temporal & 0.7497 & 0.8727 & 0.8848 & 0.8579 \\
Spatial~+~Temporal & \textbf{0.7643} & \textbf{0.8870} &	\textbf{0.9016} & \textbf{0.8726} \\
         \multicolumn{5}{c}{}\\
        \end{tabular}}} \hspace{3mm}
        \subfloat[\textbf{Study on Difference Operators.} 
        `Difference operator' refers to the distance metric applied in difference calculation.
        \label{table:operator}]{
        \scalebox{0.9}{
			\tablestyle{2pt}{1.05}
			\begin{tabular}{x{34}|x{25}x{25}x{25}x{25}}
         Operators & 0.05 & 0.25 & 0.5 & Average \\
         \shline
         Manhattan & 0.7632 & 0.8870 & \textbf{0.9024} & 0.8725 \\
        Euclidean  & \textbf{0.7643} & \textbf{0.8870} & 0.9016 & \textbf{0.8726} \\
        Chebyshev & 0.7540 & 0.8789 & 0.8931 & 0.8643 \\
        Cosine  & 0.7611 & 0.8834 &	0.8982 & 0.8691 \\
         \multicolumn{5}{c}{}\\
        \end{tabular}}} \hspace{3mm}
          \subfloat[\textbf{Study on Aggregation Methods of Progressive Attention}. `Avg' refers to direct temporal averaging operation without attention.
        \label{table:progressiveattn}]{
        \scalebox{0.9}{
			\tablestyle{2pt}{1.05}
			\begin{tabular}{x{40}|x{25}x{25}x{25}x{25}}
         Aggregation & 0.05 & 0.25 & 0.5 & Average \\
         \shline
         Avg & 0.7498 & 0.8685 & 0.8804 & 0.8543 \\
Intra & 0.7588 & 0.8793 &	0.8922 & 0.8650  \\
Cross & 0.7590 & 0.8770 &	0.8894 & 0.8631 \\
Intra~+~Cross & \textbf{0.7643} & \textbf{0.8870} & \textbf{0.9016} & \textbf{0.8726} \\
         \multicolumn{5}{c}{}\\
        \end{tabular}}} 
    \caption{\textbf{Ablation Studies on Kinetics-GEBD dataset,} measured by F1 score at different Rel.Dis. thresholds.}
\vspace{-4mm}
\label{tab:exploration}
\end{table*}

\subsection{Training}
\noindent\textbf{Balanced Sampler.}
GEBD is a binary classification task, and non-boundary frames are far more than boundary frames ($r$:1). Following~\cite{Shou_2021_ICCV}, we leverage a balanced sampler. Due to slowness prior in videos~\cite{DBLP:journals/pami/ZhangT12}, features of consecutive non-boundary frames change at a very slow speed. Hence, we apply a sparse sampling strategy on non-boundary frames, namely select one out of sequential $r$ non-boundary frames randomly and sample all boundary frames.

\noindent\textbf{Loss Function.}
The balanced sampler relieves the imbalance of positive and negative samples. Therefore, we simply define the binary classification loss $\mathcal{L}_{bc}$ as:
\vspace{-2mm}
\begin{equation}
    \mathcal{L}_{bc} = -\frac{1}{N}\sum_{\eta=1}^{N}
    (\hat{p}_\eta \log p_{\eta} + (1-\hat{p}_{\eta})\log(1-p_{\eta})),
\label{Class_loss}
\end{equation}
where $p_{\eta}$ is the binary classification probability and $N$ is the total number of training samples. $\hat{p}_{\eta}$ is 1 if the sample is marked as a boundary, and otherwise 0.

\subsection{Inference}
\noindent\textbf{Linear Fusion of Logits.} 
After progressive attention, $\mathbf{q}_{A}^{''}$ and $\mathbf{q}_{D}^{''}$ are separately passed into two independent $fc$ layers to generate logits $l_{A}$ and $l_{D}$. With a learnable parameter $\alpha$, we perform a linear fusion of logits: $l = \alpha * l_{A} + (1 - \alpha) * l_{D}$. Softmax function is applied on the final logit $l$ to get the boundary probability $p$.

\noindent\textbf{Efficient Post-processing Scheme.} 
Repeating the above process of predicting the boundary probability of one frame, we obtain the boundary confidence sequence of the whole video. To select the final boundary predictions of the video, we apply an efficient post-processing scheme on the sequence. In detail, a boundary frame should satisfy the following two requirements: (1) The boundary probability of the frame is greater than a set threshold $\theta$ (\eg, 0.5). (2) Its boundary probability is the maximum within a pre-defined range (\eg, [-5, 5]). Since our post-processing scheme is free of time-consuming pairwise IoU calculation, it only takes about 0.0003 seconds per video (5.302s for all 18,813 videos) on one Nvidia V100 machine.

\section{Experiments}
\subsection{Dataset and Setup}
\noindent\textbf{Kinetics-GEBD.} Kinetics-GEBD dataset~\cite{Shou_2021_ICCV} consists of 60,000 videos randomly selected from Kinetics-400. Among them, 18,794 training videos and 17,725 testing videos are randomly selected from Kinetics-400 training set. Kinetics-GEBD validation set contains all 18,813 videos in Kinetics-400 validation set. The ratio of training,
validation and testing sets is nearly 1:1:1. Since temporal annotations of testing set are not available, we train on the training set and evaluate with the validation set. 

\noindent\textbf{TAPOS.} TAPOS dataset~\cite{DBLP:conf/cvpr/ShaoZDL20} contains Olympics sports videos with 21 actions. There are 13,094 training action instances and 1,790 validation action instances. Following~\cite{Shou_2021_ICCV}, we re-purpose TAPOS for GEBD task by trimming each action instance with its action label hidden.

\noindent\textbf{Evaluation Protocol.}
To evaluate the results of generic event boundary detection task, we calculate F1 score and use the Relative Distance (Rel.Dis.) measurement~\cite{Shou_2021_ICCV}. Rel.Dis. is the relative distance between predictions and ground truths, divided by the length of the corresponding video. Given a threshold, a prediction is determined to be true if Rel.Dis. is smaller than or equal to the threshold, otherwise false. In experiments, we follow~\cite{Shou_2021_ICCV} to report F1 score with Rel.Dis. threshold set [0.05 : 0.05 : 0.5].

\noindent\textbf{Implementation Details.}
In practice, we select one frame out of every 3 consecutive frames, namely the stride of boundary evaluation is 3. To predict the confidence of the current frame, we take a $T\times s$ ($T=2 \times w + 1$) clip as the input, where $w$ is 5 and $s$ is 6. Following~\cite{Shou_2021_ICCV}, our model is built on ImageNet-pretrained ResNet-50 backbone and trained \emph{end-to-end}. $m$ and $n$ of multi-level feature bank are set to 3. $\omega$ of progressive attention is set to 5. To train DDM-Net, we employ Adam as the optimizer. The batch size is set to 32 and the learning rate is set to 1e-5.

\subsection{Main Results}
We fairly compare our DDM-Net with state-of-the-art methods on the validation set of Kinetics-GEBD and TAPOS. As a result, our method outperforms the state-of-the-art methods by a large margin at all Rel.Dis. thresholds.

\noindent\textbf{Kinetics-GEBD.} Table~\ref{table:GEBD} illustrates the performance of different methods on Kinetics-GEBD
validation set. It can be seen that DDM-Net remarkably outperforms other methods on F1 score, demonstrating the effectiveness of dense differences and sophisticated aggregation. Especially, DDM-Net achieves significant improvements from 62.5\% to 76.4\% at the most strict threshold (Rel.Dis.=0.05). A boost of nearly 14 percent proves the boundary predictions of DDM-Net are the most precise. Furthermore, combined with a more powerful backbone CSN~\cite{DBLP:conf/iccv/TranWFT19}, our DDM-Net can be superior to winner solutions of \href{https://sites.google.com/view/loveucvpr21}{LOVEU Challenge@CVPR 2021}, as shown in Table~\ref{table:loveu}. It is worth noting that this result is obtained without bells and whistles (\eg, model ensemble, audio data and human-object detector) of winner solutions.

\noindent\textbf{TAPOS.} The comparison results of the state-of-the-art GEBD methods on TAPOS are summarized in Table~\ref{table:TAPOS}. Since DDM-Net is able to learn complex semantics and distinguish subtle changes between sub-actions, it obtains state-of-the-art performance on TAPOS, increasing F1 score@0.05 from 52.2\% to 60.4\%. The result proves that our model can not only achieve accurate generic event boundary detection (Kinetics-GEBD), but also precisely detect boundaries between fine-grained sub-actions (TAPOS).

\subsection{Ablation Study}

\noindent\textbf{Study on Different Representations.}
We analyze our proposed DDM by experimenting with different representations, which is shown in Table~\ref{table:representation}. \emph{First}, we compare the performance of single representation. Our proposed DDM outperforms RGB, optical flow and RGB differences by a large margin, especially under strict settings (Rel.Dis. = 0.05). \emph{Second}, we find that DDM brings the greatest improvements when combined with RGB features, which proves the complementarity of dense differences and RGB features. Furthermore, as DDM is calculated on-the-fly, our method is free of training two separate networks and thus is more efficient than previous two-stream methods.

\noindent\textbf{Study on Temporal Views.}
In Section~\ref{section:multilevel}, we argue that boundary frames have stronger correlations with their adjacent snippets than faraway frames. To prove our point, we experiment with clips of different temporal views. As illustrated in Table~\ref{table:view}, the temporal view grows with the increase of stride $s$, but the performance decreases in row 3 and 4. Particularly, when $s$ is increased to 30, namely we classify the current frame with uniformly sampled frames through the whole video (about 300 frames), the performance witnesses a significant drop. Hence, it is unnecessary to introduce global temporal contexts in GEBD task. In addition, if the temporal view is fixed, using more frames ($w=15$ instead of $w=5$) to calculate denser differences can slightly improve the performance (row 2 and row 5).

\noindent\textbf{Study on Multi-Level Feature Bank.}
In order to model various motion patterns of diverse event boundaries, we first build a multi-level feature bank, then calculate dense differences upon it. As demonstrated in Table~\ref{table:multilevel}, both multi-level spatial and temporal features outperform the result with single-level features. Moreover, a combination of multi-level spatial and temporal features (firstly generate multi-level spatial features, then yield multi-level temporal features upon spatial features of each layer) can further lead to a remarkable performance advance, confirming the overall contributions of multi-level feature bank.

\noindent\textbf{Study on Difference Operators.}
In Table~\ref{table:operator}, Euclidean distance, Manhattan distance and Cosine distance are respectively employed to measure the feature difference. We observe that our method obtains close performance. Therefore, performance of DDM-Net is robust to the choice of difference operators. In contrast, our model witnesses a drop with Chebyshev distance. Since Chebyshev distance measures the max discrepancy among all channels, it cannot reflect the overall differences between two frames.
  
\begin{figure}[t]
\begin{center}
\includegraphics[scale=0.2]{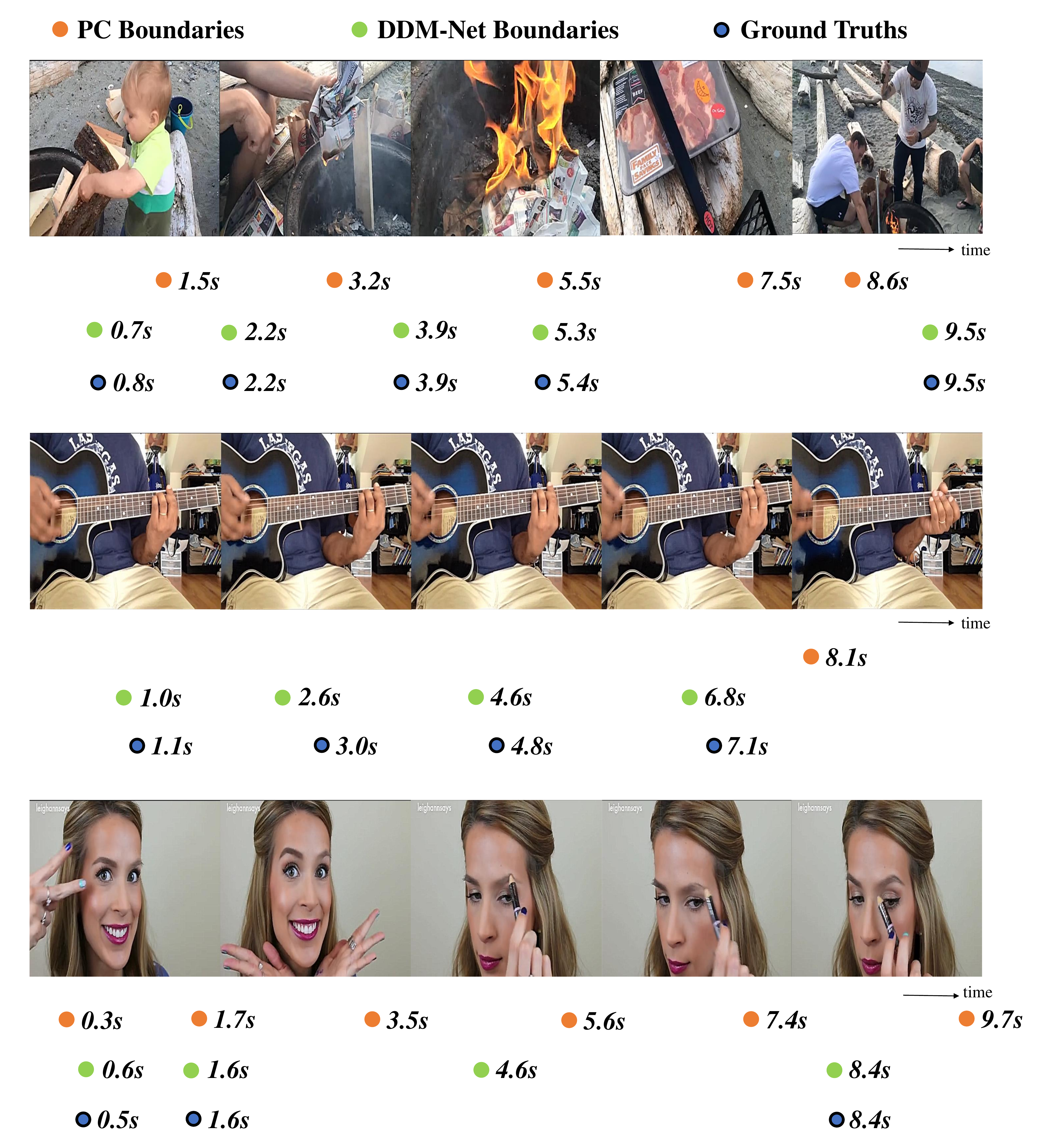}
\vspace{-2mm}
\caption{\textbf{Qualitative results} and comparisons of PC, DDM-Net and ground truths on Kinetics-GEBD dataset.}
\label{fig:qualitative}
\end{center}
\vspace{-8mm}
\end{figure}

\noindent\textbf{Study on Aggregation Methods of Progressive Attention.}
In Table~\ref{table:progressiveattn}, we study two aggregation methods of progressive attention (map-squeezed attention is required to align the shape and cannot be removed). Intra-modal attention mainly focuses on aggregating and enhancing key intra-modal features with learnable queries $\mathbf{q}$. Compared with the overall feature sequence $H$, refined queries $\mathbf{q}^{'}$ contain key patterns and less noise, which can explain the performance gain of row 2. If only cross-modal attention is leveraged, overall RGB features $A$ and DDM features $D$ directly query and guide each other through joint feature learning across modalities, leading to the improvement of row 3. Moreover, instead of $H$, cross-modal aggregation of refined key features $\mathbf{q}^{'}$ can further boost the performance (row 4).

\subsection{Qualitative Results}
Figure~\ref{fig:qualitative} displays qualitative results of our method, including different types of event boundaries. The first example is a video of several shot changes. DDM-Net precisely perceives temporal variations and hits every boundary instance, while predictions of PC are not accurate. Case in the second row is more challenging, as only the position of the left hand changes. DDM-Net is able to model complex semantics and distinguish subtle action changes, therefore it makes accurate predictions. In contrast, PC misses all the ground truths. The last example is a combination of shot changes and action changes. Since our method is more robust to temporal noise (camera jitter), it predicts fewer false positives. In summary, thanks to multi-level DDM and progressive attention, our method is able to precisely perceive temporal changes and understand complicated semantics, hence it has shown advantages in many different cases.

\section{Conclusion}
In this paper, we have presented a modular framework for the task of generic event boundary detection (GEBD). To perceive diverse temporal variations and learn complex semantics of generic event boundaries, our method progressively attends to multi-level dense difference maps (DDM). Thanks to holistic temporal modeling and joint feature learning across modalities, our DDM-Net outperforms the previous state-of-the-art methods by a large margin on Kinetics-GEBD and TAPOS benchmark. In addition, our method is better than winner solutions of LOVEU Challenge@CVPR 2021, further demonstrating the efficacy of DDM-Net. As for \textbf{limitations}, large-scale GEBD benchmarks of untrimmed videos are expected to further validate our method in future work.

\vspace{-2mm}
\paragraph {\bf Acknowledgements.} {\small This work is supported by National Natural Science Foundation of China  (No.62076119, No.61921006),  Program for Innovative Talents and Entrepreneur in Jiangsu Province, and Collaborative Innovation Center of Novel Software Technology and Industrialization.}

\newpage
\appendix
\setcounter{equation}{0}
\setcounter{section}{0}
\setcounter{subsection}{0}
\setcounter{table}{0}
\setcounter{figure}{0}

\section*{Appendix}
\renewcommand\thesection{\Alph{section}}
\renewcommand\thetable{\Alph{table}}
\renewcommand\thefigure{\Alph{figure}}
\renewcommand\theequation{\Alph{equation}}

\section{Additional Ablation Study}

\noindent\textbf{Study on Multi-Level Spatial Architecture.} To further analyze the design of our multi-level spatial feature bank, we perform comparisons of different multi-level spatial architectures in Table~\ref{table:fpn}. To be specific, we respectively employ feature pyramid network (FPN~\cite{DBLP:conf/cvpr/LinDGHHB17}) and pyramidal feature hierarchy (SSD~\cite{DBLP:conf/eccv/LiuAESRFB16}) as the multi-level spatial feature extractor of DDM-Net, and compare the performance. Different from prior knowledge in object detection, DDM-Net witnesses an inferior performance when combined with FPN. We analyze that lateral and top-down connection layers of FPN are trained without explicit spatial location supervisions (\eg, bounding boxes) in GEBD task~\cite{Shou_2021_ICCV}, thus leading to insufficient training of those layers and overall performance degradation.

\begin{table}[!h]
\begin{center}
\vspace{-1mm}
\scalebox{0.82}{
\begin{tabular}{c|cccc}
\toprule[1pt]
Spatial architecture  & 0.05 & 0.25 & 0.5 & Average \\ 
\hline
Feature pyramid network & 0.7511 & 0.8697 & 0.8815 & 0.8557 \\
Pyramidal feature hierarchy & \textbf{0.7643} & \textbf{0.8870} &	\textbf{0.9016} & \textbf{0.8726} \\
\bottomrule[1pt]
\end{tabular}
}
\vspace{-2mm}
\caption{Study on multi-level spatial architecture on Kinetics-GEBD, measured by F1 score at different Rel.Dis. thresholds.}
\vspace{-4mm}
\label{table:fpn}
\end{center}
\end{table}

\noindent\textbf{Study on the Number of Attention Layers.} We experiment on the number of attention layers of intra-modal attention module and cross-modal attention module, and display the results in Table~\ref{table:num_attn}. DDM-Net achieves the best performance with 6 intra-modal attention layers and 6 cross-modal attention layers. With the increase of the number of attention layers, the performance gain of increasing layers decreases.

\begin{table}[!h]
\begin{center}
\vspace{-1mm}
\scalebox{0.9}{
\begin{tabular}{cc|cccc}
\toprule[1pt]
Intra & Cross & 0.05 & 0.25 & 0.5 & Average \\ 
\hline
1 & 1 & 0.7584 & 0.8774 & 0.8895 & 0.8633 \\
3 & 3 & 0.7622 & 0.8841 & 0.8983 & 0.8698 \\
6 & 6 & \textbf{0.7643} & \textbf{0.8870} &	\textbf{0.9016} & \textbf{0.8726} \\
\bottomrule[1pt]
\end{tabular}
}
\vspace{-2mm}
\caption{Study on the number of attention layers on Kinetics-GEBD, measured by F1 score at different Rel.Dis. thresholds.}
\vspace{-4mm}
\label{table:num_attn}
\end{center}
\end{table}

\noindent\textbf{Study on the Number of Learnable Queries $\omega$.} The number of learnable queries $\omega$ influences the performance of DDM-Net, as demonstrated in Table~\ref{table:num_query}. Too few queries ($\omega=1$) are not enough to capture all patterns, while too many queries ($\omega=10$) lead to redundant intra-modal features. In experiments, we observe that DDM-Net reaches the best performance when $\omega$ is set to 5. 

\begin{table}[!h]
\begin{center}
\vspace{-1mm}
\scalebox{0.9}{
\begin{tabular}{c|cccc}
\toprule[1pt]
$\omega$ & 0.05 & 0.25 & 0.5 & Average \\ 
\hline
1 & 0.7579 & 0.8763 & 0.8884 & 0.8622 \\
3 & 0.7614 & 0.8853 & 0.9004 & 0.8709 \\
5 & \textbf{0.7643} & \textbf{0.8870} &	\textbf{0.9016} & \textbf{0.8726} \\
10  & 0.7592 & 0.8765 & 0.8888 & 0.8626 \\
\bottomrule[1pt]
\end{tabular}
}
\vspace{-2mm}
\caption{Study on the number of learnable queries $\omega$ on Kinetics-GEBD, measured by F1 score at different Rel.Dis. thresholds.}
\vspace{-6mm}
\label{table:num_query}
\end{center}
\end{table}

\noindent\textbf{Study on Positional Embeddings of Cross-Modal Attention.} We conduct ablations on positional embeddings of cross-modal attention module, as shown in Table~\ref{table:pe}. Adding positional embeddings in the cross-modal attention module harms the performance, hence we analyze that cross-modality feature aggregation should be directly operated on raw features. It is worth noting that learnable positional embeddings of intra-modal attention module cannot be removed, as they are employed to localize key features of clips, similar to positional embeddings in previous detection tasks (\eg, to localize objects in object detection, to localize actions in temporal action detection). 

\begin{table}[!h]
\begin{center}
\vspace{-1mm}
\scalebox{0.85}{
\begin{tabular}{c|cccc}
\toprule[1pt]
Aggregation & 0.05 & 0.25 & 0.5 & Average \\ 
\hline
cross~w/~PE & 0.7510 & 0.8720 & 0.8841 &	0.8576 \\
cross~w/o~PE & 0.7590 & 0.8770 & 0.8894 & 0.8631 \\
\hline
intra~+~~cross~w/~PE & 0.7597 & 0.8801 & 0.8931 & 0.8659 \\
intra~+~cross~w/o~PE & \textbf{0.7643} & \textbf{0.8870} &	\textbf{0.9016} & \textbf{0.8726} \\
\bottomrule[1pt]
\end{tabular}
}
\end{center}
\vspace{-4mm}
\begin{tablenotes}
\footnotesize
\item * PE: positional embedding, w/: with, w/o: without.
\end{tablenotes}
\vspace{-2mm}
\caption{Study on positional embedding of cross-modal attention module on Kinetics-GEBD, measured by F1 score at different Rel.Dis. thresholds.}
\label{table:pe}
\vspace{-2mm}
\end{table}

\noindent\textbf{Study on Balanced Sampler.} Since boundaries and non-boundaries are extremely imbalanced (about 1:6), we follow ~\cite{Shou_2021_ICCV} to exploit the same balanced sampler. As shown in Table~\ref{table:sampler}, the performance of balanced sampler is better.

\begin{table}[!h]
\begin{center}
\scalebox{0.82}{
\begin{tabular}{c|cccc}
\toprule[1pt]
Sampler  & 0.05 & 0.25 & 0.5 & Average \\ 
\hline
Plain sampler & 0.7456 & 0.8784 & 0.8932 & 0.8627 \\
Balanced sampler & \textbf{0.7643} & \textbf{0.8870} &	\textbf{0.9016} & \textbf{0.8726} \\
\bottomrule[1pt]
\end{tabular}
}
\caption{Study on balanced sampler on Kinetics-GEBD, measured by F1 score at different Rel.Dis. thresholds.}
\vspace{-4mm}
\label{table:sampler}
\end{center}
\end{table}

\section{More Results}
\noindent\textbf{More comparisons of different representations.} Due to the limited space, we present the best performance of each representation in Table 4a of the main paper. In this section, we display the complete result of each representation and perform more comparisons of different representations. \emph{First,} only DDM obtains the best performance when stride $s$ is set to 6, indicating that compared with other representations, DDM can take advantage of larger temporal contexts. \textit{Second}, if we fairly compare the representations under the same setting $s=3$, DDM still outperforms other representations, demonstrating the effectiveness of dense motion representation in GEBD task. \textit{Third}, in Table~\ref{table:pairwise}, pairwise flow and RGB differences are superior to consecutive (non-pairwise) flow and RGB differences at the most strict threshold (Rel.Dis. = 0.05), demonstrating the effectiveness of pairwise calculation in GEBD task.

\begin{table}[!h]
\begin{center}
\vspace{-1mm}
\scalebox{0.85}{
\begin{tabular}{c|cc|cccc}
\toprule[1pt]
Representation & $w$ & $s$ & 0.05 & 0.25 & 0.5 & Average \\ 
\hline
RGB & 5 & 3 & \textbf{0.6793} & \textbf{0.8589} &	\textbf{0.8772} & \textbf{0.8375} \\
RGB & 5 & 6 & 0.6118 & 0.8462 &	0.8772 & 0.8180 \\
\hline
 Flow & 5 & 3 & \textbf{0.6625} & \textbf{0.8045} & \textbf{0.8206} & \textbf{0.7877} \\
Flow & 5 & 6 & 0.6091 & 0.7703 & 0.7975 & 0.7530
\\
\hline
RGB diff & 5 & 3 & \textbf{0.7272} & \textbf{0.8591} & \textbf{0.8753} & \textbf{0.8440} \\
RGB diff & 5 & 6 & 0.6638 & 0.8629 & 0.8876 & 0.8399 \\
\hline
DDM & 5 & 3 & 0.7476 & 0.8688 & 0.8813 & 0.8544 \\
DDM & 5 & 6 & \textbf{0.7512} &	\textbf{0.8738} & \textbf{0.8861} & \textbf{0.8591}\\
\bottomrule[1pt]
\end{tabular}
}
\label{table:more_representation}
\end{center}
\vspace{-2mm}
\begin{tablenotes}
\footnotesize
\item * $w$ and $s$ are defined in the main paper.
\end{tablenotes}
\vspace{-2mm}
\caption{More comparisons of different representations on Kinetics-GEBD, measured by F1 score at different Rel.Dis. thresholds.}
\end{table}

\begin{table}[!h]
\begin{center}
\vspace{-1mm}
\scalebox{0.85}{
\begin{tabular}{c|cccc}
\toprule[1pt]
Representation & 0.05 & 0.25 & 0.5 & Average \\ 
\hline
Flow & 0.6625 & \textbf{0.8045} & \textbf{0.8206} & \textbf{0.7877} \\
Pairwise Flow & \textbf{0.7012} & 0.7910 & 0.7998 & 0.7806 \\
\hline
RGB diff & 0.7272 & 0.8591 & 0.8753 & 0.8440 \\
Pairwise RGB diff & \textbf{0.7311} & \textbf{0.8617} & \textbf{0.8753} & \textbf{0.8461} \\
\bottomrule[1pt]
\end{tabular}
}
\vspace{-2mm}
\caption{Comparisons of pairwise and non-pairwise flow and RGB differences on the validation set of Kinetics-GEBD, measured by F1 score at different Rel.Dis. thresholds.}
\vspace{-4mm}
\label{table:pairwise}
\end{center}
\end{table}

\noindent\textbf{DDM-Net with CSN backbone.} Owing to the limited space of the main paper, we report the performance of DDM-Net on testing set when it is combined with IG-65M~\cite{DBLP:conf/cvpr/GhadiyaramTM19} pretrained CSN~\cite{DBLP:conf/iccv/TranWFT19} backbone network. To validate our method, we further perform ablations on the validation set (annotations of the testing set are not available, entries to the testing server are limited).  
In Table~\ref{table:csnval}, we observe that DDM-Net can still increase the performance of powerful CSN representations by nearly 2 percent, from 79.3\% to 81.3\%.

\begin{table}[!h]
\begin{center}
\vspace{-1mm}
\scalebox{0.85}{
\begin{tabular}{c|cccc}
\toprule[1pt]
Model & 0.05 & 0.25 & 0.5 & Average \\ 
\hline
CSN + FC & 0.7933 & 0.8954 & 0.9074 & 0.8834 \\
CSN + DDM-Net  & \textbf{0.8128} & \textbf{0.9077} & \textbf{0.9218} & \textbf{0.8972} \\
\bottomrule[1pt]
\end{tabular}
}
\vspace{-2mm}
\caption{Performance of DDM-Net with CSN backbone on the validation set of Kinetics-GEBD, measured by F1 score at different Rel.Dis. thresholds.}
\vspace{-4mm}
\label{table:csnval}
\end{center}
\end{table}

\begin{figure}[!h]
 \centering
 \subfloat[Sparse and dense motion representations of boundary examples.]{\includegraphics[width=3.3in]{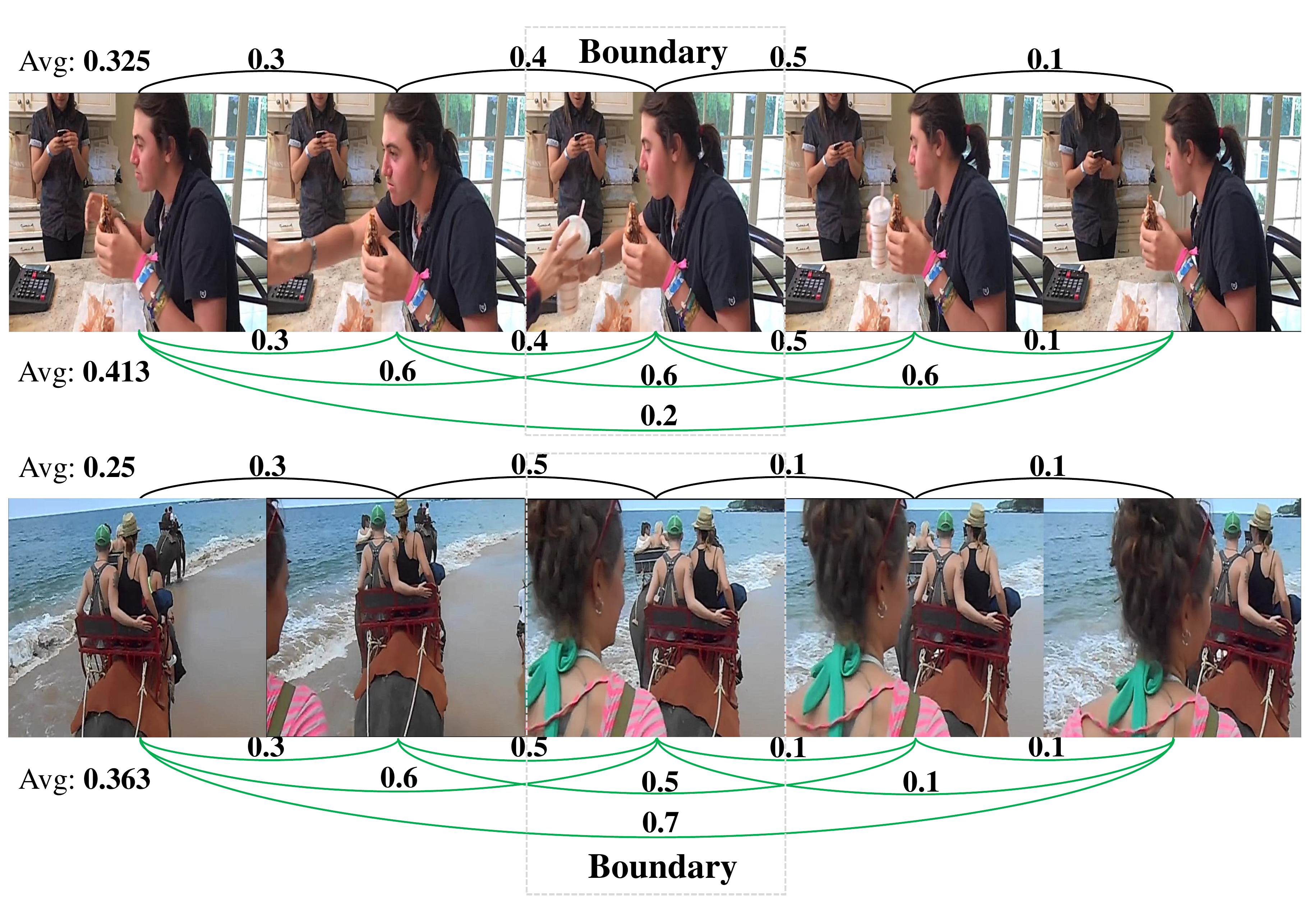}} 
 \label{s}\\
 \subfloat[Sparse and dense motion representations of non-boundary examples.]{\includegraphics[width=3.3in]{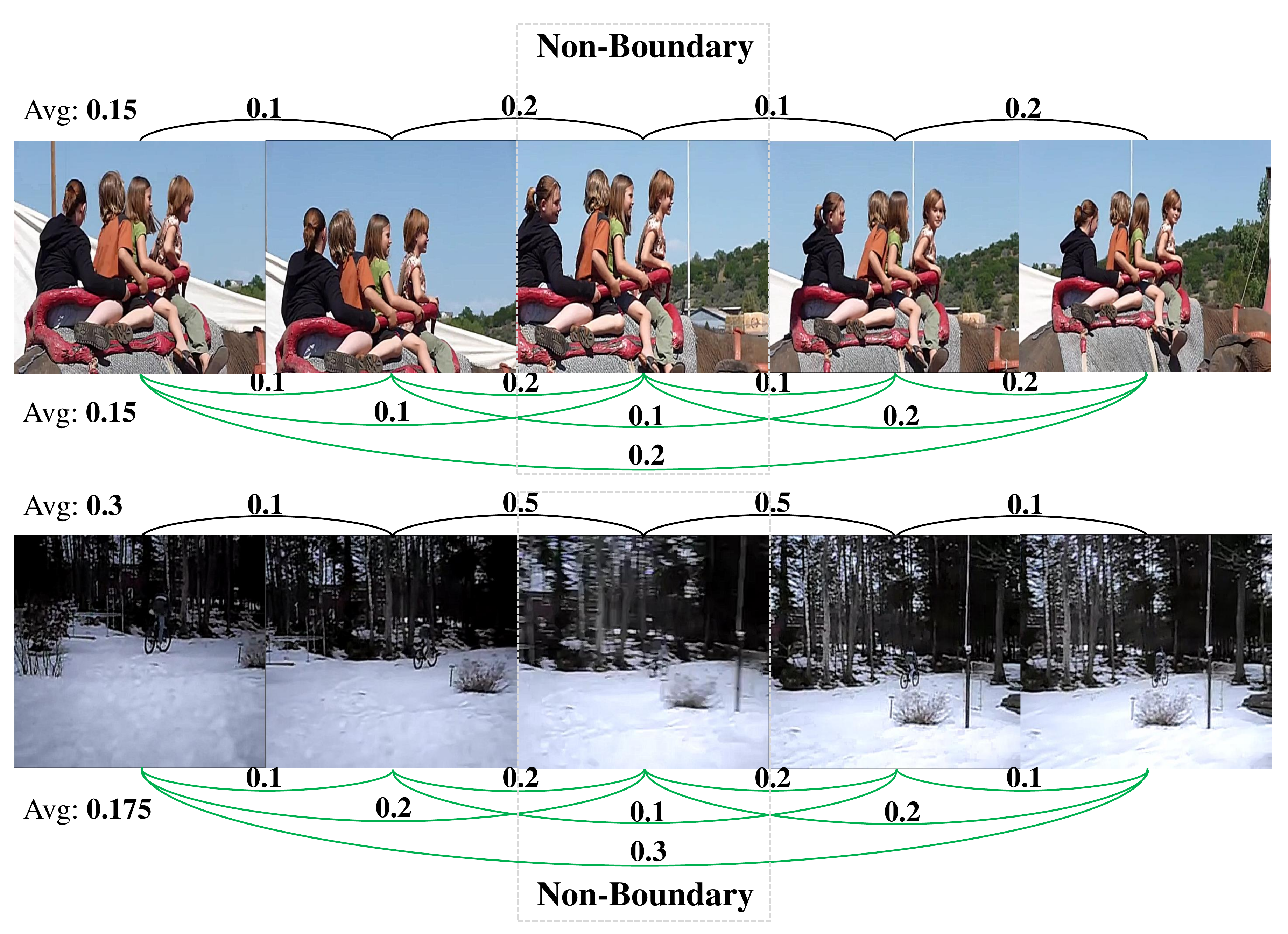}}
 \caption{More comparisons of sparse motion representation (black lines, optical flow) and dense motion representation (green lines, some are omitted for clarity, dense feature differences). Numbers on lines indicate the magnitude of motion between two frames. Dense motion representation provides more holistic temporal cues to better distinguish boundaries and non-boundaries.}
 \vspace{-6mm}
 \label{fig:motion}
\end{figure}

\noindent\textbf{Time analysis.} Average time cost of DDM-Net (0.123s) is close to ~\cite{Shou_2021_ICCV} (0.066s). We calculate pairwise differences of frames in a sliding window ($T$ frames, the sliding stride is $s$) rather than the whole video of $E$ frames. Hence, the run-time complexity of DDM in a video is $O((E/s)\times T\times T)$ instead of $O(E\times E)$. In practice, $T$ could be much smaller than $E$ (\eg, $E$ = 300, $T$ = 11, $s$ = 3). Furthermore, computations can be re-used for subsequent frames since sliding windows are overlapped.

\noindent\textbf{Complete Results of Precision, Recall and F1 score.} Complete results of precision, recall and F1 score are presented in Table~\ref{table:full_kinGEBD} and Table~\ref{table:full_TAPOS}. It is noteworthy that several methods (SceneDetect~\cite{castellano2014PySceneDetect}, PA~\cite{Shou_2021_ICCV}) achieve high precision yet very low recall, while several methods (TCN~\cite{DBLP:conf/eccv/LeaRVH16}) obtain high recall yet low precision. The first type of methods (high precision yet very low recall) focus on salient boundaries (\eg, shot changes) and miss other event boundaries, while the second type of methods (high recall yet low precision) make as many predictions as possible and recall many false positives. As a result, both of them do not achieve superior F1 score.

As for competitive methods~\cite{kang2021winning, hong2021generic, sun2021generic} of LOVEU challenge, they do not present results on the standard validation set of Kinetics-GEBD for fair comparisons. Therefore, we compare DDM-Net with them on the testing set of Kinetics-GEBD in Table 3 of main paper (evaluated on the server of the challenge).

\begin{figure*}
\begin{center}
\includegraphics[scale=0.45]{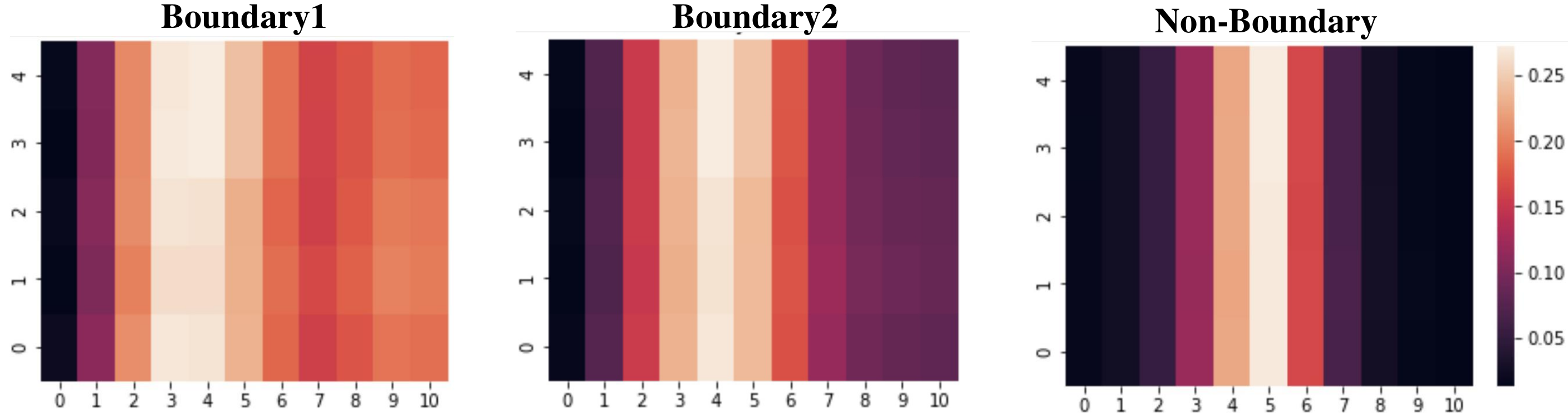}
\end{center}
\vspace{-2mm}
   \caption{Visualization of cross-attention weight maps in the intra-modal attention module, averaged among multiple heads of the last cross-attention layer. The y-axis is event queries and the x-axis represents timestamps of features. The color represents the magnitude of weight, as the weight becomes larger from black to white. Best viewed in color.}
\label{fig:intra_cross}
\vspace{-2mm}
\end{figure*}

\begin{figure}
\begin{center}
\includegraphics[scale=0.15]{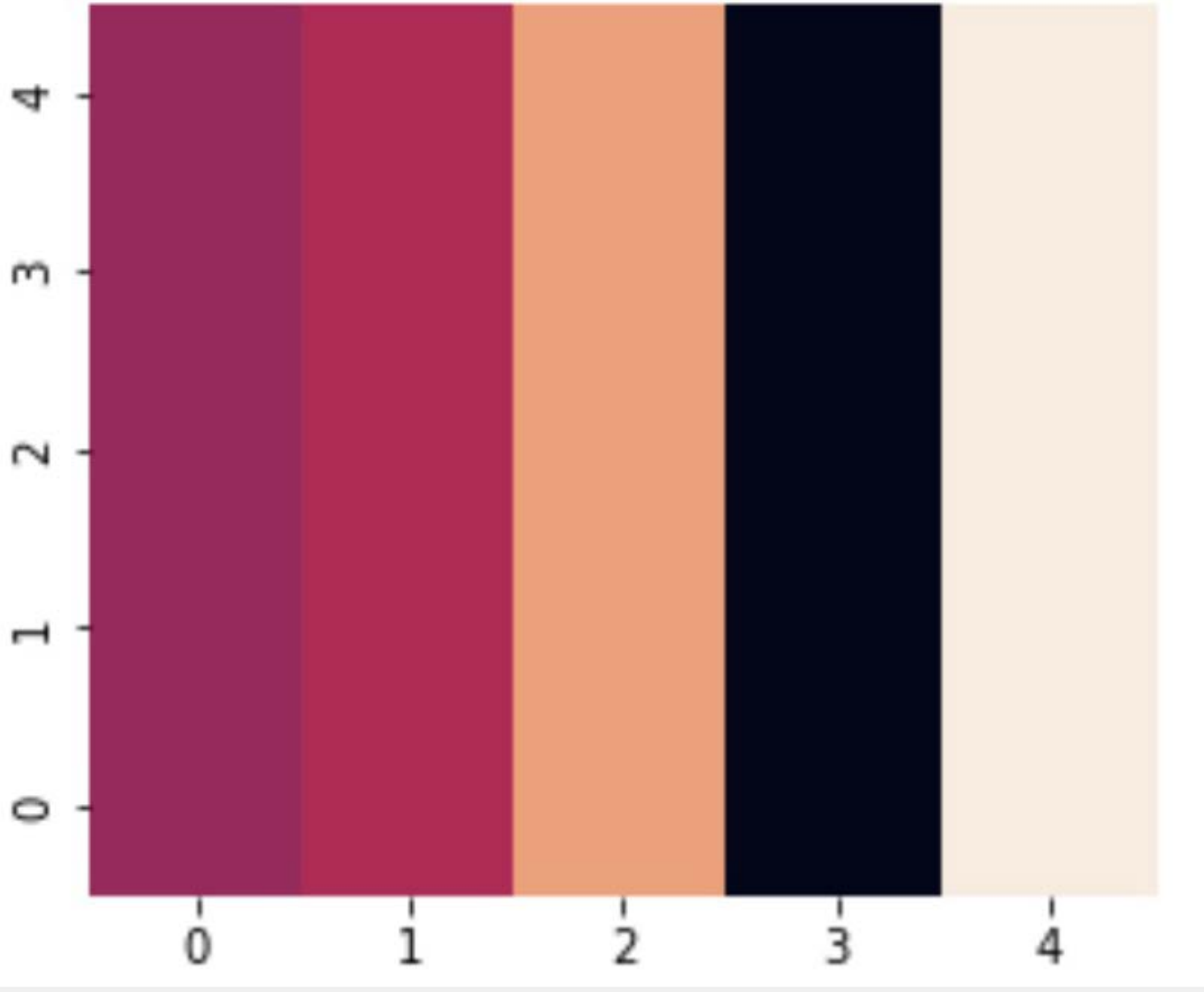}
\end{center}
\vspace{-2mm}
   \caption{Visualization of cross-attention weight maps in the cross-modal attention module, averaged among multiple heads of the last cross-attention layer. The y-axis and the x-axis represent event queries. Best viewed in color.}
\label{fig:cross_cross}
\vspace{-2mm}
\end{figure}

\section{Visualization}
\noindent\textbf{More Comparisons of Sparse and Dense Motion Representation.} We add more comparisons of sparse and dense motion representation, as displayed in Figure~\ref{fig:motion}. Figure~\ref{fig:motion}(a) illustrates two kinds of boundaries. The example in the first row is an `event A$\rightarrow$event B$\rightarrow$event A' boundary (content of the boundary frame is different from other frames), while the second one is an `event A$\rightarrow$event B' boundary (the boundary frame is one frame of event A or event B). In both cases, the average magnitude of dense motion representation is larger than sparse motion representation, enabling the model to detect boundaries more easily. Figure~\ref{fig:motion}(b) also displays two kinds of non-boundaries. The first one is a non-boundary without large temporal changes, while the example in the second row is a non-boundary with temporal noise (camera blur). As our proposed DDM is calculated upon multi-level features instead of raw frames, it is more robust to noise than optical flow. Hence, the average magnitude of dense motion representation is smaller than sparse motion representation. Comparing the average magnitude in Figure~\ref{fig:motion}(a) and Figure~\ref{fig:motion}(b), we observe that holistic temporal clues of dense motion representation enable the model to better distinguish boundaries and non-boundaries.   

\noindent\textbf{Visualization of Progressive Attention Module.} To further explore the effects of Progressive Attention Module, we present attention weight maps of intra-modal attention module and cross-modal attention module.
In both Figure~\ref{fig:intra_cross} and Figure~\ref{fig:cross_cross}, differences between \textit{columns} are significant, indicating that features of several moments or queries are enhanced. Figure~\ref{fig:intra_cross} displays $T\times \omega$ cross-attention weight maps of intra-modal attention module. We observe that weight maps of non-boundaries are similar, as shown in the right subfigure. Since there are no obvious temporal changes in non-boundaries, attention weights of queries approximately follow Gaussian distribution (the weight decreases from the center frame to both sides). In contrast, weight maps of boundaries are diverse (left 2 subfigures), where queries attend to moments where temporal changes happen. Weight map of the last cross-attention layer in cross-modal attention module is shown in Figure~\ref{fig:cross_cross}, where features of several queries are enhanced under the cross-modality guidance.

\begin{figure*}
\begin{center}
\includegraphics[scale=0.28]{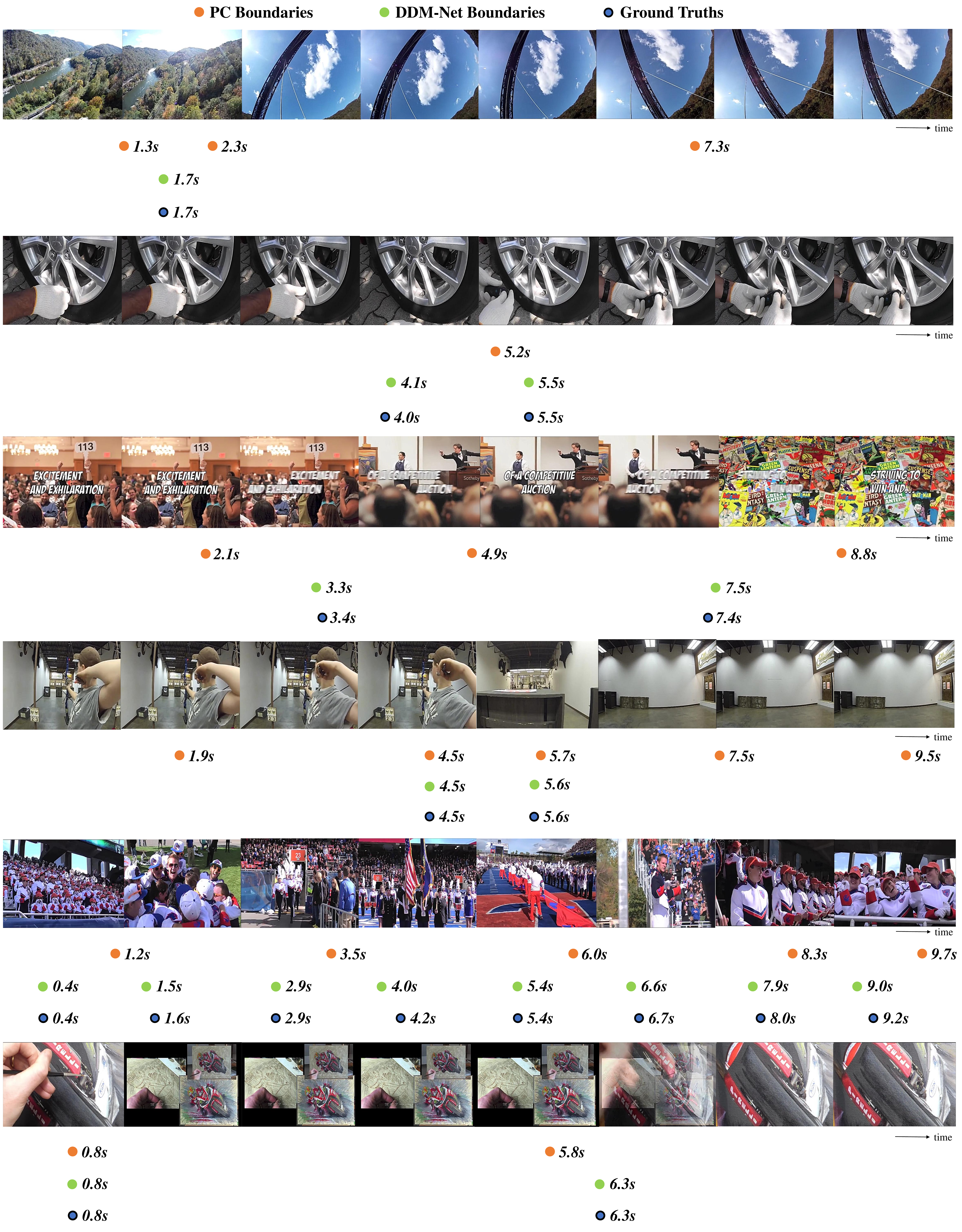}
\end{center}
\vspace{-2mm}
   \caption{More qualitative results and comparisons of PC, DDM-Net and ground truths on Kinetics-GEBD dataset.}
\label{fig:more_qualitative}
\vspace{-2mm}
\end{figure*}

\noindent\textbf{More Qualitative Results.} We add more qualitative results to demonstrate the effectiveness of our proposed DDM-Net, as illustrated in Figure~\ref{fig:more_qualitative}. According to those examples, we conclude that DDM-Net predicts fewer false positives (high precision) and hits more ground truths (high recall) than PC~\cite{Shou_2021_ICCV}, thus obtains a superior F1 score. 

\section{More Implementation Details}
\noindent\textbf{Complete Loss Function.} We only present one classification loss function in the main paper because of the limited space. In practice, the loss function is the sum of 3 binary classification losses:
\begin{equation}
\begin{aligned}
    \mathcal{L}_{bc} &= \frac{1}{N}\sum_{\eta=1}^{N}(\mathcal{L}_{fu, \eta} + \mathcal{L}_{A, \eta} + \mathcal{L}_{D, \eta}),\\
    \mathcal{L}_{fu, \eta} &= -(\hat{p}_\eta\log p_{fu, \eta} + (1-\hat{p}_{\eta})\log(1-p_{fu, \eta})),\\
    \mathcal{L}_{A, \eta} &= -(\hat{p}_\eta\log p_{A, \eta} + (1-\hat{p}_{\eta})\log(1-p_{A, \eta})),\\
    \mathcal{L}_{D, \eta} &= -(\hat{p}_\eta\log p_{D, \eta} + (1-\hat{p}_{\eta})\log(1-p_{D, \eta})),\\
\end{aligned}
\label{complete_loss}
\end{equation}
where $p_{fu, \eta}$, $p_{A, \eta}$ and $p_{D, \eta}$ are binary classification probabilities of fusion logit $l$, appearance logit $l_A$ and difference logit $l_D$ of the sample. $N$ is the total number of training samples. $\hat{p}_{\eta}$ is 1 if the sample is marked as a boundary, and otherwise 0.  

\begin{table*}[t]
\small
\begin{center}
\begin{tabular}{ccccccccccccc}
\multicolumn{13}{c}{(a) Precision}                                                                                                                                                                                                                                                                 \\ \hline
\multicolumn{2}{c|}{Rel.Dis. threshold}                                            & 0.05           & 0.1            & 0.15           & 0.2            & 0.25           & 0.3            & 0.35           & 0.4            & 0.45           & \multicolumn{1}{c|}{0.5}            & avg            \\ \hline
\multicolumn{1}{c|}{\multirow{3}{*}{Unsuper.}} & \multicolumn{1}{c|}{SceneDetect~\cite{castellano2014PySceneDetect}}  & 0.731          & 0.792          & 0.819          & 0.837          & 0.847          & 0.856          & 0.862          & 0.867          & 0.870          & \multicolumn{1}{c|}{0.872}          & 0.835          \\
\multicolumn{1}{c|}{}                          & \multicolumn{1}{c|}{PA - Random~\cite{Shou_2021_ICCV}}    & 0.737                     & \multicolumn{1}{c}{0.884} & \multicolumn{1}{c}{0.933} & \multicolumn{1}{c}{0.956} & \multicolumn{1}{c}{0.968} & \multicolumn{1}{c}{0.975} & \multicolumn{1}{c}{0.979} & \multicolumn{1}{c}{0.981} & \multicolumn{1}{c}{0.984} & \multicolumn{1}{c|}{0.986}          & \multicolumn{1}{c}{0.938}          \\
\multicolumn{1}{c|}{}                          
& \multicolumn{1}{c|}{PA~\cite{Shou_2021_ICCV}}           & 0.836          & 0.944          & 0.965          & 0.973          & 0.978          & 0.980          & 0.983          & 0.985          & 0.986          & \multicolumn{1}{c|}{0.989}          & 0.962          \\ \hline

\multicolumn{1}{c|}{\multirow{6}{*}{Super.}}   & \multicolumn{1}{c|}{BMN~\cite{DBLP:conf/iccv/LinLLDW19}}          & 0.128          & 0.141          & 0.148          & 0.152          & 0.156          & 0.159          & 0.162          & 0.164          & 0.165          & \multicolumn{1}{c|}{0.167}          & 0.154          \\
\multicolumn{1}{c|}{}                          & \multicolumn{1}{c|}{BMN-StartEnd~\cite{DBLP:conf/iccv/LinLLDW19}} & 0.396          & 0.479          & 0.509          & 0.525          & 0.534          & 0.540          & 0.544          & 0.547          & 0.549          & \multicolumn{1}{c|}{0.551}          & 0.517          \\
\multicolumn{1}{c|}{}                          & \multicolumn{1}{c|}{TCN-TAPOS~\cite{DBLP:conf/eccv/LeaRVH16}}    & 0.518          & 0.622          & 0.665          & 0.690          & 0.706          & 0.718          & 0.727          & 0.733          & 0.738          & \multicolumn{1}{c|}{0.743}          & 0.686          \\
\multicolumn{1}{c|}{}                          & \multicolumn{1}{c|}{TCN~\cite{DBLP:conf/eccv/LeaRVH16}}          & 0.461          & 0.519          & 0.538          & 0.547          & 0.553          & 0.557          & 0.559          & 0.561          & 0.563          & \multicolumn{1}{c|}{0.564}          & 0.542          \\
\multicolumn{1}{c|}{}                          & \multicolumn{1}{c|}{PC~\cite{Shou_2021_ICCV}}           & 0.624          & 0.753          & 0.794          & 0.816          & 0.828          & 0.836          & 0.841          & 0.844          & 0.846          & \multicolumn{1}{c|}{0.849}          & 0.803          \\ 
\cline{2-13}
\multicolumn{1}{c|}{}                          & \multicolumn{1}{c|}{DDM-Net} 
& 0.732          & 0.812          & 0.836          & 0.849          & 0.856          & 0.860          & 0.863          & 0.865          & 0.867          & \multicolumn{1}{c|}{0.869}          & 0.841          \\ \hline
                                               &                                   &                &                &                &                &                &                &                &                &                &                                     &                \\
\multicolumn{13}{c}{(b) Recall}                                                                                                                                                                                                                                                                    \\ \hline
\multicolumn{2}{c|}{Rel.Dis. threshold}                                            & 0.05           & 0.1            & 0.15           & 0.2            & 0.25           & 0.3            & 0.35           & 0.4            & 0.45           & \multicolumn{1}{c|}{0.5}            & avg            \\ \hline
\multicolumn{1}{c|}{\multirow{3}{*}{Unsuper.}} & \multicolumn{1}{c|}{SceneDetect~\cite{castellano2014PySceneDetect}}  & 0.170          & 0.185          & 0.192          & 0.197          & 0.200          & 0.202          & 0.204          & 0.206          & 0.207          & \multicolumn{1}{c|}{0.207}          & 0.197          \\
\multicolumn{1}{c|}{}                          
& \multicolumn{1}{c|}{PA - Random~\cite{Shou_2021_ICCV}}    & 0.218          & 0.289          & 0.326          & 0.350          & 0.364          & 0.374          & 0.381          & 0.386          & 0.389          & \multicolumn{1}{c|}{0.393}          & 0.347          \\

\multicolumn{1}{c|}{}                          
& \multicolumn{1}{c|}{PA~\cite{Shou_2021_ICCV}}           & 0.259          & 0.329         & 0.355         & 0.368          & 0.377          & 0.382         & 0.386          & 0.390          & 0.392          & \multicolumn{1}{c|}{0.395}          & 0.363          \\ \hline

\multicolumn{1}{c|}{\multirow{6}{*}{Super.}}   & \multicolumn{1}{c|}{BMN~\cite{DBLP:conf/iccv/LinLLDW19}}          & 0.338          & 0.369          & 0.385          & 0.397          & 0.407          & 0.414          & 0.420          & 0.426          & 0.430          & \multicolumn{1}{c|}{0.434}          & 0.402          \\
\multicolumn{1}{c|}{}                          & \multicolumn{1}{c|}{BMN-StartEnd~\cite{DBLP:conf/iccv/LinLLDW19}} & 0.648          & 0.766          & 0.817          & 0.846          & 0.864          & 0.876          & 0.885          & 0.892          & 0.897          & \multicolumn{1}{c|}{0.900}          & 0.839          \\
\multicolumn{1}{c|}{}                          & \multicolumn{1}{c|}{TCN-TAPOS~\cite{DBLP:conf/eccv/LeaRVH16}}    & 0.420          & 0.508          & 0.550          & 0.576          & 0.594          & 0.609          & 0.619          & 0.627          & 0.633          & \multicolumn{1}{c|}{0.639}          & 0.577          \\
\multicolumn{1}{c|}{}                          & \multicolumn{1}{c|}{TCN~\cite{DBLP:conf/eccv/LeaRVH16}}          & 0.811          & 0.894          & 0.923          & 0.938          & 0.947          & 0.952          & 0.956          & 0.959          & 0.961          & \multicolumn{1}{c|}{0.963}          & 0.930          \\
\multicolumn{1}{c|}{}                          & \multicolumn{1}{c|}{PC~\cite{Shou_2021_ICCV}}           & 0.626          & 0.764          & 0.814          & 0.843          & 0.859          & 0.871          & 0.879          & 0.885          & 0.889          & \multicolumn{1}{c|}{0.892}          & 0.832          \\
\cline{2-13}
\multicolumn{1}{c|}{}                          & \multicolumn{1}{c|}{DDM-Net} 
& 0.800          & 0.875          & 0.899          & 0.912          & 0.920          & 0.926          & 0.930          & 0.933          & 0.935          & \multicolumn{1}{c|}{0.937}          & 0.907          \\ \hline
                                               &                                   &                &                &                &                &                &                &                &                &                &                                     &                \\
\multicolumn{13}{c}{(c) F1}                                                                                                                                                                                                                                                                        \\ \toprule[1pt]
\multicolumn{2}{c|}{Rel.Dis. threshold}                                            & 0.05           & 0.1            & 0.15           & 0.2            & 0.25           & 0.3            & 0.35           & 0.4            & 0.45           & \multicolumn{1}{c|}{0.5}            & avg            \\ 
\hline
\multicolumn{1}{c|}{\multirow{3}{*}{Unsuper.}} & \multicolumn{1}{c|}{SceneDetect~\cite{castellano2014PySceneDetect}}  & 0.275 & 0.300          & 0.312          & 0.319          & 0.324          & 0.327          & 0.330         & 0.332          & 0.334          & \multicolumn{1}{c|}{0.335}          & 0.318          \\
\multicolumn{1}{c|}{}                          & \multicolumn{1}{c|}{PA - Random~\cite{Shou_2021_ICCV}}    & 0.336          & 0.435          & 0.484         & 0.512          & 0.529          & 0.541          & 0.548          & 0.554          & 0.558          & \multicolumn{1}{c|}{0.561}          & 0.506         \\
\multicolumn{1}{c|}{}                          & \multicolumn{1}{c|}{PA~\cite{Shou_2021_ICCV}}      & 0.396          & 0.488 & 0.520 & 0.534 & 0.544 & 0.550 & 0.555 & 0.558 & 0.561 & \multicolumn{1}{c|}{0.564} & 0.527 \\
\hline
\multicolumn{1}{c|}{\multirow{6}{*}{Super.}}   & \multicolumn{1}{c|}{BMN~\cite{DBLP:conf/iccv/LinLLDW19}}          & 0.186          & 0.204          & 0.213          & 0.220          & 0.226          & 0.230          & 0.233          & 0.237          & 0.239          & \multicolumn{1}{c|}{0.241}          & 0.223          \\
\multicolumn{1}{c|}{}                          & \multicolumn{1}{c|}{BMN-StartEnd~\cite{DBLP:conf/iccv/LinLLDW19}} & 0.491          & 0.589          & 0.627          & 0.648          & 0.660          & 0.668          & 0.674          & 0.678          & 0.681          & \multicolumn{1}{c|}{0.683}          & 0.640          \\
\multicolumn{1}{c|}{}                          & \multicolumn{1}{c|}{TCN-TAPOS~\cite{DBLP:conf/eccv/LeaRVH16}}    & 0.464          & 0.560          & 0.602          & 0.628          & 0.645          & 0.659          & 0.669          & 0.676          & 0.682          & \multicolumn{1}{c|}{0.687}          & 0.627          \\
\multicolumn{1}{c|}{}                          & \multicolumn{1}{c|}{TCN~\cite{DBLP:conf/eccv/LeaRVH16}}          & 0.588          & 0.657          & 0.679          & 0.691          & 0.698          & 0.703          & 0.706          & 0.708          & 0.710          & \multicolumn{1}{c|}{0.712}          & 0.685          \\
\multicolumn{1}{c|}{}                          & \multicolumn{1}{c|}{PC~\cite{Shou_2021_ICCV}}            & 0.625  & 0.758  & 0.804  & 0.829  & 0.844  & 0.853  & 0.859  & 0.864  & 0.867  & \multicolumn{1}{c|}{0.870}          &   0.817 \\ 
\cline{2-13}
\multicolumn{1}{c|}{}                          & \multicolumn{1}{c|}{DDM-Net}            & \textbf{0.764}  & \textbf{0.843}  & \textbf{0.866}  & \textbf{0.880}  & \textbf{0.887}  & \textbf{0.892}  & \textbf{0.895}  & \textbf{0.898}  & \textbf{0.900}  & \multicolumn{1}{c|}{\textbf{0.902}}          &   \textbf{0.873}
\\ \bottomrule[1pt]
\end{tabular}
\end{center}
\caption{Precision, Recall and F1 score of state-of-the-art GEBD methods on Kinetics-GEBD.}
\label{table:full_kinGEBD}
\end{table*}

\begin{table*}[t]
\small
\begin{tabular}{cccllllllllll}
\multicolumn{13}{c}{(a) Precision}                                                                                                                                                                                                                                                                                                                                                                              \\ \hline
\multicolumn{2}{c|}{Rel.Dis. threshold}                                           & 0.05                      & \multicolumn{1}{c}{0.1}   & \multicolumn{1}{c}{0.15}  & \multicolumn{1}{c}{0.2}   & \multicolumn{1}{c}{0.25}  & \multicolumn{1}{c}{0.3}   & \multicolumn{1}{c}{0.35}  & \multicolumn{1}{c}{0.4}   & \multicolumn{1}{c}{0.45}  & \multicolumn{1}{c|}{0.5}            & \multicolumn{1}{c}{avg}   \\ \hline
\multicolumn{1}{c|}{\multirow{3}{*}{Unsuper.}} & \multicolumn{1}{c|}{SceneDetect} & 0.391                     & \multicolumn{1}{c}{0.506} & \multicolumn{1}{c}{0.532} & \multicolumn{1}{c}{0.576} & \multicolumn{1}{c}{0.596} & \multicolumn{1}{c}{0.608} & \multicolumn{1}{c}{0.621} & \multicolumn{1}{c}{0.628} & \multicolumn{1}{c}{0.641} & \multicolumn{1}{c|}{0.647}          & \multicolumn{1}{c}{0.575} \\
\multicolumn{1}{c|}{}                          & 

\multicolumn{1}{c|}{PA - Random~\cite{Shou_2021_ICCV}}   & 0.206                     & \multicolumn{1}{c}{0.304} & \multicolumn{1}{c}{0.356} & \multicolumn{1}{c}{0.404} & \multicolumn{1}{c}{0.432} & \multicolumn{1}{c}{0.452} & \multicolumn{1}{c}{0.466} & \multicolumn{1}{c}{0.481} & \multicolumn{1}{c}{0.491} & \multicolumn{1}{c|}{0.500}          & \multicolumn{1}{c}{0.409} \\

\multicolumn{1}{c|}{}                          &
\multicolumn{1}{c|}{PA~\cite{Shou_2021_ICCV}}          & 0.470                     & \multicolumn{1}{c}{0.599} & \multicolumn{1}{c}{0.662} & \multicolumn{1}{c}{0.708} & \multicolumn{1}{c}{0.740} & \multicolumn{1}{c}{0.755} & \multicolumn{1}{c}{0.771} & \multicolumn{1}{c}{0.784} & \multicolumn{1}{c}{0.795} & \multicolumn{1}{c|}{0.801}          & \multicolumn{1}{c}{0.708} \\ \hline

\multicolumn{1}{c|}{\multirow{6}{*}{Super.}}   & \multicolumn{1}{c|}{ISBA~\cite{DBLP:conf/cvpr/DingX18}}        & 0.119                     & \multicolumn{1}{c}{0.185} & \multicolumn{1}{c}{0.230} & \multicolumn{1}{c}{0.268} & \multicolumn{1}{c}{0.301} & \multicolumn{1}{c}{0.329} & \multicolumn{1}{c}{0.356} & \multicolumn{1}{c}{0.379} & \multicolumn{1}{c}{0.392} & \multicolumn{1}{c|}{0.405}          & \multicolumn{1}{c}{0.296} \\
\multicolumn{1}{c|}{}                          & \multicolumn{1}{c|}{TCN~\cite{DBLP:conf/eccv/LeaRVH16}}         & 0.140                     & \multicolumn{1}{c}{0.187} & \multicolumn{1}{c}{0.200} & \multicolumn{1}{c}{0.204} & \multicolumn{1}{c}{0.207} & \multicolumn{1}{c}{0.208} & \multicolumn{1}{c}{0.210} & \multicolumn{1}{c}{0.211} & \multicolumn{1}{c}{0.211} & \multicolumn{1}{c|}{0.211}          & \multicolumn{1}{c}{0.199} \\
\multicolumn{1}{c|}{}                          & \multicolumn{1}{c|}{CTM~\cite{DBLP:conf/eccv/HuangFN16}}         & 0.154                     & \multicolumn{1}{c}{0.197} & \multicolumn{1}{c}{0.212} & \multicolumn{1}{c}{0.221} & \multicolumn{1}{c}{0.228} & \multicolumn{1}{c}{0.233} & \multicolumn{1}{c}{0.237} & \multicolumn{1}{c}{0.242} & \multicolumn{1}{c}{0.244} & \multicolumn{1}{c|}{0.245}          & \multicolumn{1}{c}{0.221} \\
\multicolumn{1}{c|}{}                          & \multicolumn{1}{c|}{TransParser~\cite{DBLP:conf/cvpr/ShaoZDL20}} & 0.230                     & \multicolumn{1}{c}{0.302} & \multicolumn{1}{c}{0.345} & \multicolumn{1}{c}{0.377} & \multicolumn{1}{c}{0.398} & \multicolumn{1}{c}{0.410} & \multicolumn{1}{c}{0.420} & \multicolumn{1}{c}{0.427} & \multicolumn{1}{c}{0.432} & \multicolumn{1}{c|}{0.437}          & \multicolumn{1}{c}{0.378} \\
\multicolumn{1}{c|}{}                          & \multicolumn{1}{c|}{PC~\cite{Shou_2021_ICCV}}          & 0.650                     & \multicolumn{1}{c}{0.741} & \multicolumn{1}{c}{0.782} & \multicolumn{1}{c}{0.805} & \multicolumn{1}{c}{0.821} & \multicolumn{1}{c}{0.829} & \multicolumn{1}{c}{0.836} & \multicolumn{1}{c}{0.842} & \multicolumn{1}{c}{0.846} & \multicolumn{1}{c|}{0.851}          & \multicolumn{1}{c}{0.800} \\ 
\cline{2-13}
\multicolumn{1}{c|}{}                          & \multicolumn{1}{c|}{DDM-Net} 
& 0.591          & 0.667          & 0.700          & 0.720          & 0.732          & 0.737          & 0.741          & 0.744          & 0.748          & \multicolumn{1}{c|}{0.751}          & 0.713          \\ \hline

\multicolumn{1}{l}{}                           & \multicolumn{1}{l}{}             & \multicolumn{1}{l}{}      &                           &                           &                           &                           &                           &                           &                           &                           &                                     &                           \\
\multicolumn{13}{c}{(b) Recall}                                                                                                                                                                                                                                                                                                                                                                                 \\ \hline
\multicolumn{2}{c|}{Rel.Dis. threshold}                                           & 0.05                      & \multicolumn{1}{c}{0.1}   & \multicolumn{1}{c}{0.15}  & \multicolumn{1}{c}{0.2}   & \multicolumn{1}{c}{0.25}  & \multicolumn{1}{c}{0.3}   & \multicolumn{1}{c}{0.35}  & \multicolumn{1}{c}{0.4}   & \multicolumn{1}{c}{0.45}  & \multicolumn{1}{c|}{0.5}            & \multicolumn{1}{c}{avg}   \\ \hline
\multicolumn{1}{c|}{\multirow{3}{*}{Unsuper.}} & \multicolumn{1}{c|}{SceneDetect} & \multicolumn{1}{l}{0.018} & 0.023                     & 0.025                     & 0.027                     & 0.028                     & 0.028                     & 0.029                     & 0.029                     & 0.030                     & \multicolumn{1}{l|}{0.030}          & 0.027                     \\
\multicolumn{1}{c|}{}                          
& \multicolumn{1}{c|}{PA - Random~\cite{Shou_2021_ICCV}}   & \multicolumn{1}{l}{0.128} & 0.189                     & 0.221                     & 0.252                     & 0.269                     & 0.281                     & 0.290                     & 0.299                     & 0.305                     & \multicolumn{1}{l|}{0.311}          & 0.255                     \\

\multicolumn{1}{c|}{}                          & 
\multicolumn{1}{c|}{PA~\cite{Shou_2021_ICCV}}          & \multicolumn{1}{l}{0.292} & 0.372                    & 0.412                     & 0.440                     & 0.460                     & 0.470                     & 0.480                    & 0.488                     & 0.494                     & \multicolumn{1}{l|}{0.498}          & 0.441                     \\ \hline

\multicolumn{1}{c|}{\multirow{6}{*}{Super.}}   & \multicolumn{1}{c|}{ISBA~\cite{DBLP:conf/cvpr/DingX18}}        & \multicolumn{1}{l}{0.095} & 0.158                     & 0.225                     & 0.263                     & 0.296                     & 0.323                     & 0.340                     & 0.360                     & 0.373                     & \multicolumn{1}{l|}{0.386}          & 0.282                     \\
\multicolumn{1}{c|}{}                          & \multicolumn{1}{c|}{TCN~\cite{DBLP:conf/eccv/LeaRVH16}}         & \multicolumn{1}{l}{0.757} & 0.940                     & 0.974                     & 0.985                     & 0.989                     & 0.990                     & 0.994                     & 0.994                     & 0.994                     & \multicolumn{1}{l|}{0.994}          & 0.961                     \\
\multicolumn{1}{c|}{}                          & \multicolumn{1}{c|}{CTM~\cite{DBLP:conf/eccv/HuangFN16}}         & \multicolumn{1}{l}{0.596} & 0.752                     & 0.811                     & 0.843                     & 0.860                     & 0.875                     & 0.886                     & 0.894                     & 0.898                     & \multicolumn{1}{l|}{0.901}          & 0.831                     \\
\multicolumn{1}{c|}{}                          & \multicolumn{1}{c|}{TransParser~\cite{DBLP:conf/cvpr/ShaoZDL20}} & \multicolumn{1}{l}{0.386} & 0.516                     & 0.590                     & 0.642                     & 0.673                     & 0.689                     & 0.705                     & 0.714                     & 0.721                     & \multicolumn{1}{l|}{0.726}          & 0.636                     \\
\multicolumn{1}{c|}{}                          & \multicolumn{1}{c|}{PC~\cite{Shou_2021_ICCV}}          & \multicolumn{1}{l}{0.436} & 0.497                     & 0.525                     & 0.541                     & 0.551                     & 0.556                     & 0.561                     & 0.565                     & 0.568                     & \multicolumn{1}{l|}{0.572}          & 0.537                     \\
\cline{2-13}
\multicolumn{1}{c|}{}                          & \multicolumn{1}{c|}{DDM-Net} 
& 0.617          & 0.695          & 0.730          & 0.751          & 0.764          & 0.769          & 0.774          & 0.777          & 0.780          & \multicolumn{1}{c|}{0.783}          & 0.744          \\ \hline

\multicolumn{1}{l}{}                           & \multicolumn{1}{l}{}             & \multicolumn{1}{l}{}      &                           &                           &                           &                           &                           &                           &                           &                           &                                     &                           \\
\multicolumn{13}{c}{(c) F1}                                                                                                                                                                                                                                                                                                                                                                                     \\ \toprule[1pt]
\multicolumn{2}{c|}{Rel.Dis. threshold}                                           & 0.05                      & \multicolumn{1}{c}{0.1}   & \multicolumn{1}{c}{0.15}  & \multicolumn{1}{c}{0.2}   & \multicolumn{1}{c}{0.25}  & \multicolumn{1}{c}{0.3}   & \multicolumn{1}{c}{0.35}  & \multicolumn{1}{c}{0.4}   & \multicolumn{1}{c}{0.45}  & \multicolumn{1}{c|}{0.5}            & \multicolumn{1}{c}{avg}   \\ \hline
\multicolumn{1}{c|}{\multirow{3}{*}{Unsuper.}} & \multicolumn{1}{c|}{SceneDetect} & \multicolumn{1}{l}{0.035} & 0.045                     & 0.047                     & 0.051                     & 0.053                     & 0.054                     & 0.055                     & 0.056                     & 0.057                     & \multicolumn{1}{l|}{0.058}          & 0.051                     \\
\multicolumn{1}{c|}{}                          & \multicolumn{1}{c|}{PA - Random~\cite{Shou_2021_ICCV}}   & 0.158          & 0.233          & 0.273          & 0.310          & 0.331          & 0.347          & 0.357          & 0.369          & 0.376          & \multicolumn{1}{c|}{0.384}          &    0.314                       \\
\multicolumn{1}{c|}{}                          &
\multicolumn{1}{c|}{PA~\cite{Shou_2021_ICCV}}      & 0.360 & 0.459 & 0.507 & 0.543 & 0.567 & 0.579 & 0.592 & 0.601 & 0.609 & \multicolumn{1}{c|}{0.615} & 0.543 \\
\hline
\multicolumn{1}{c|}{\multirow{6}{*}{Super.}}   & \multicolumn{1}{c|}{ISBA~\cite{DBLP:conf/cvpr/DingX18}}        & \multicolumn{1}{l}{0.106} & 0.170                     & 0.227                     & 0.265                     & 0.298                     & 0.326                     & 0.348                     & 0.369                     & 0.382                     & \multicolumn{1}{l|}{0.396}          & 0.302                     \\
\multicolumn{1}{c|}{}                          & \multicolumn{1}{c|}{TCN~\cite{DBLP:conf/eccv/LeaRVH16}}         & 0.237                     & 0.312                     & 0.331                     & 0.339                     & 0.342                     & 0.344                     & 0.347                     & 0.348                     & 0.348                     & \multicolumn{1}{l|}{0.348}          & 0.330                     \\
\multicolumn{1}{c|}{}                          & \multicolumn{1}{c|}{CTM~\cite{DBLP:conf/eccv/HuangFN16}}         & 0.244                     & 0.312                     & 0.336                     & 0.351                     & 0.361                     & 0.369                     & 0.374                     & 0.381                     & 0.383                     & \multicolumn{1}{l|}{0.385}          & 0.350                     \\
\multicolumn{1}{c|}{}                          & \multicolumn{1}{c|}{TransParser~\cite{DBLP:conf/cvpr/ShaoZDL20}} & 0.289                     & 0.381                     & 0.435                     & 0.475                     & 0.500                     & 0.514                     & 0.527                     & 0.534                     & 0.540                     & \multicolumn{1}{l|}{0.545}          & 0.474                     \\
\multicolumn{1}{c|}{}                          & \multicolumn{1}{c|}{PC~\cite{Shou_2021_ICCV}}    & 0.522          & 0.595          & 0.628          & 0.646          & 0.659       & 0.665          & 0.671          & 0.676          & 0.679          & \multicolumn{1}{c|}{0.683}          & 0.642        \\ 
\cline{2-13}
\multicolumn{1}{c|}{}                          & \multicolumn{1}{c|}{DDM-Net}            & \textbf{0.604}  & \textbf{0.681}  & \textbf{0.715}  & \textbf{0.735}  & \textbf{0.747}  & \textbf{0.753}  & \textbf{0.757}  & \textbf{0.760}  & \textbf{0.763}  & \multicolumn{1}{c|}{\textbf{0.767}}          &   \textbf{0.728}
\\ \bottomrule[1pt]
\end{tabular}                                                       
\vspace{1.5em}
\caption{Precision, Recall and F1 score of state-of-the-art GEBD methods on TAPOS.}
\label{table:full_TAPOS}
\end{table*}

\noindent\textbf{Detailed Formulas of Progressive Attention Module.} Due to the limited space of the main paper, we define $\mathbf{q}$, $\mathbf{k}$ and $\mathbf{v}$ of intra-modal attention module and cross-modal attention module respectively. In this section, we present the complete inference process of attention modules with detailed formulas. First, we review Attention and Multi-Head Attention mechanism~\cite{DBLP:conf/nips/VaswaniSPUJGKP17},
\begin{equation}
\begin{aligned}
    \text{Attn}(\mathbf{q},\mathbf{k},\mathbf{v}) &= \text{softmax}(\frac{\mathbf{q}\mathbf{k}^\intercal}{\sqrt{d_k}})\mathbf{v},\\
    \text{MHA}(\mathbf{q},\mathbf{k},\mathbf{v}) &= \text{Concat}(\text{head}_1, \cdots, \text{head}_\mathbf{H})\mathbf{W}^O, \\
     \text{head}_i &= \text{Attn}(\mathbf{q}\mathbf{W}_i^\mathbf{q}, \mathbf{k}\mathbf{W}_i^\mathbf{k}, \mathbf{v}\mathbf{W}_i^\mathbf{v}),\\
\end{aligned}
\label{attn_mechanism}
\end{equation}
where $d_k$ is the dimension of features and $\mathbf{W}$ is the learnable projection matrix to transform the feature. Then, the calculation of each layer in intra-modal attention module and cross-modal attention module can be formulated as:
\begin{equation}
\begin{aligned}
    &\mathbf{q'} = \text{LN}(\mathbf{q} + \text{MHA}(\mathbf{q}, \mathbf{q}, \mathbf{q})),\\
    &\mathbf{q''} = \text{LN}(\mathbf{q'} + \text{MHA}(\mathbf{q'}, \mathbf{k}, \mathbf{v})),\\
     &\text{Output} = \text{LN}(\mathbf{q''} + \text{FFN}(\mathbf{q''})),\\
\end{aligned}
\end{equation}
where $\text{LN}$ and $\text{FFN}$ 
denote layer normalization and feed forward network.

{\small
\bibliographystyle{ieee_fullname}
\bibliography{egbib}

\begin{thebibliography}{10}\itemsep=-1pt

\bibitem{DBLP:conf/cvpr/00010BT0GZ18}
Peter Anderson, Xiaodong He, Chris Buehler, Damien Teney, Mark Johnson, Stephen
  Gould, and Lei Zhang.
\newblock Bottom-up and top-down attention for image captioning and visual
  question answering.
\newblock In {\em {CVPR}}, pages 6077--6086. Computer Vision Foundation /
  {IEEE} Computer Society, 2018.

\bibitem{DBLP:conf/caip/BaraldiGC15}
Lorenzo Baraldi, Costantino Grana, and Rita Cucchiara.
\newblock Shot and scene detection via hierarchical clustering for re-using
  broadcast video.
\newblock In {\em {CAIP} {(1)}}, volume 9256 of {\em Lecture Notes in Computer
  Science}, pages 801--811. Springer, 2015.

\bibitem{DBLP:conf/bmvc/BuchEGFN17}
Shyamal Buch, Victor Escorcia, Bernard Ghanem, Li Fei{-}Fei, and Juan~Carlos
  Niebles.
\newblock End-to-end, single-stream temporal action detection in untrimmed
  videos.
\newblock In {\em {BMVC}}. {BMVA} Press, 2017.

\bibitem{DBLP:conf/cvpr/CarreiraZ17}
Jo{\~{a}}o Carreira and Andrew Zisserman.
\newblock Quo vadis, action recognition? {A} new model and the kinetics
  dataset.
\newblock In {\em {CVPR}}, pages 4724--4733. {IEEE} Computer Society, 2017.

\bibitem{castellano2014PySceneDetect}
Brandon Castellano.
\newblock {PySceneDetect}: an intelligent scene cut detection and video
  splitting tool.
\newblock \url{https://github.com/Breakthrough/PySceneDetect}, 2014.

\bibitem{chen2021endtoend}
Yi-Wen Chen, Yi-Hsuan Tsai, and Ming-Hsuan Yang.
\newblock End-to-end multi-modal video temporal grounding.
\newblock In A. Beygelzimer, Y. Dauphin, P. Liang, and J.~Wortman Vaughan,
  editors, {\em Advances in Neural Information Processing Systems}, 2021.

\bibitem{DBLP:conf/cvpr/DingX18}
Li Ding and Chenliang Xu.
\newblock Weakly-supervised action segmentation with iterative soft boundary
  assignment.
\newblock In {\em {CVPR}}, pages 6508--6516. Computer Vision Foundation /
  {IEEE} Computer Society, 2018.

\bibitem{DBLP:conf/cvpr/DwibediATSZ20}
Debidatta Dwibedi, Yusuf Aytar, Jonathan Tompson, Pierre Sermanet, and Andrew
  Zisserman.
\newblock Counting out time: Class agnostic video repetition counting in the
  wild.
\newblock In {\em {CVPR}}, pages 10384--10393. Computer Vision Foundation /
  {IEEE}, 2020.

\bibitem{DBLP:conf/cvpr/FeichtenhoferPZ16}
Christoph Feichtenhofer, Axel Pinz, and Andrew Zisserman.
\newblock Convolutional two-stream network fusion for video action recognition.
\newblock In {\em {CVPR}}, pages 1933--1941. {IEEE} Computer Society, 2016.

\bibitem{DBLP:conf/iccv/GaoSYN17}
Jiyang Gao, Chen Sun, Zhenheng Yang, and Ram Nevatia.
\newblock {TALL:} temporal activity localization via language query.
\newblock In {\em {ICCV}}, pages 5277--5285. {IEEE} Computer Society, 2017.

\bibitem{DBLP:conf/eccv/GaoFG18}
Ruohan Gao, Rog{\'{e}}rio~Schmidt Feris, and Kristen Grauman.
\newblock Learning to separate object sounds by watching unlabeled video.
\newblock In {\em {ECCV} {(3)}}, volume 11207 of {\em Lecture Notes in Computer
  Science}, pages 36--54. Springer, 2018.

\bibitem{DBLP:conf/cvpr/GhadiyaramTM19}
Deepti Ghadiyaram, Du Tran, and Dhruv Mahajan.
\newblock Large-scale weakly-supervised pre-training for video action
  recognition.
\newblock In {\em {CVPR}}, pages 12046--12055. Computer Vision Foundation /
  {IEEE}, 2019.

\bibitem{DBLP:conf/iccv/GongLLSMVH19}
Dong Gong, Lingqiao Liu, Vuong Le, Budhaditya Saha, Moussa~Reda Mansour, Svetha
  Venkatesh, and Anton van~den Hengel.
\newblock Memorizing normality to detect anomaly: Memory-augmented deep
  autoencoder for unsupervised anomaly detection.
\newblock In {\em {ICCV}}, pages 1705--1714. {IEEE}, 2019.

\bibitem{DBLP:conf/cbmi/Gygli18}
Michael Gygli.
\newblock Ridiculously fast shot boundary detection with fully convolutional
  neural networks.
\newblock In {\em {CBMI}}, pages 1--4. {IEEE}, 2018.

\bibitem{DBLP:conf/iccv/HendricksWSSDR17}
Lisa~Anne Hendricks, Oliver Wang, Eli Shechtman, Josef Sivic, Trevor Darrell,
  and Bryan~C. Russell.
\newblock Localizing moments in video with natural language.
\newblock In {\em {ICCV}}, pages 5804--5813. {IEEE} Computer Society, 2017.

\bibitem{hong2021generic}
Dexiang Hong, Congcong Li, Longyin Wen, Xinyao Wang, and Libo Zhang.
\newblock Generic event boundary detection challenge at cvpr 2021 technical
  report: Cascaded temporal attention network (castanet), 2021.

\bibitem{DBLP:conf/eccv/HuangFN16}
De{-}An Huang, Li Fei{-}Fei, and Juan~Carlos Niebles.
\newblock Connectionist temporal modeling for weakly supervised action
  labeling.
\newblock In {\em {ECCV} {(4)}}, volume 9908 of {\em Lecture Notes in Computer
  Science}, pages 137--153. Springer, 2016.

\bibitem{DBLP:conf/iccv/JiangWGWY19}
Boyuan Jiang, Mengmeng Wang, Weihao Gan, Wei Wu, and Junjie Yan.
\newblock {STM:} spatiotemporal and motion encoding for action recognition.
\newblock In {\em {ICCV}}, pages 2000--2009. {IEEE}, 2019.

\bibitem{DBLP:journals/pami/JunejoDLP11}
Imran~N. Junejo, Emilie Dexter, Ivan Laptev, and Patrick P{\'{e}}rez.
\newblock View-independent action recognition from temporal self-similarities.
\newblock {\em {IEEE} Trans. Pattern Anal. Mach. Intell.}, 33(1):172--185,
  2011.

\bibitem{kang2021winning}
Hyolim Kang, Jinwoo Kim, Kyungmin Kim, Taehyun Kim, and Seon~Joo Kim.
\newblock Winning the cvpr'2021 kinetics-gebd challenge: Contrastive learning
  approach, 2021.

\bibitem{DBLP:conf/eccv/LeaRVH16}
Colin Lea, Austin Reiter, Ren{\'{e}} Vidal, and Gregory~D. Hager.
\newblock Segmental spatiotemporal cnns for fine-grained action segmentation.
\newblock In {\em {ECCV} {(3)}}, volume 9907 of {\em Lecture Notes in Computer
  Science}, pages 36--52. Springer, 2016.

\bibitem{DBLP:conf/emnlp/LeiYBB18}
Jie Lei, Licheng Yu, Mohit Bansal, and Tamara~L. Berg.
\newblock {TVQA:} localized, compositional video question answering.
\newblock In {\em {EMNLP}}, pages 1369--1379. Association for Computational
  Linguistics, 2018.

\bibitem{DBLP:conf/cvpr/LiJSZKW20}
Yan Li, Bin Ji, Xintian Shi, Jianguo Zhang, Bin Kang, and Limin Wang.
\newblock {TEA:} temporal excitation and aggregation for action recognition.
\newblock In {\em {CVPR}}, pages 906--915. Computer Vision Foundation / {IEEE},
  2020.

\bibitem{DBLP:conf/cvpr/Lin0LWTWLHF21}
Chuming Lin, Chengming Xu, Donghao Luo, Yabiao Wang, Ying Tai, Chengjie Wang,
  Jilin Li, Feiyue Huang, and Yanwei Fu.
\newblock Learning salient boundary feature for anchor-free temporal action
  localization.
\newblock In {\em {CVPR}}, pages 3320--3329. Computer Vision Foundation /
  {IEEE}, 2021.

\bibitem{DBLP:conf/cvpr/LinDGHHB17}
Tsung{-}Yi Lin, Piotr Doll{\'{a}}r, Ross~B. Girshick, Kaiming He, Bharath
  Hariharan, and Serge~J. Belongie.
\newblock Feature pyramid networks for object detection.
\newblock In {\em {CVPR}}, pages 936--944. {IEEE} Computer Society, 2017.

\bibitem{DBLP:conf/iccv/LinLLDW19}
Tianwei Lin, Xiao Liu, Xin Li, Errui Ding, and Shilei Wen.
\newblock {BMN:} boundary-matching network for temporal action proposal
  generation.
\newblock In {\em {ICCV}}, pages 3888--3897. {IEEE}, 2019.

\bibitem{DBLP:conf/eccv/LinZSWY18}
Tianwei Lin, Xu Zhao, Haisheng Su, Chongjing Wang, and Ming Yang.
\newblock {BSN:} boundary sensitive network for temporal action proposal
  generation.
\newblock In {\em {ECCV} {(4)}}, volume 11208 of {\em Lecture Notes in Computer
  Science}, pages 3--21. Springer, 2018.

\bibitem{DBLP:conf/eccv/LiuAESRFB16}
Wei Liu, Dragomir Anguelov, Dumitru Erhan, Christian Szegedy, Scott~E. Reed,
  Cheng{-}Yang Fu, and Alexander~C. Berg.
\newblock {SSD:} single shot multibox detector.
\newblock In {\em {ECCV} {(1)}}, volume 9905 of {\em Lecture Notes in Computer
  Science}, pages 21--37. Springer, 2016.

\bibitem{DBLP:conf/cvpr/LiuLLG18}
Wen Liu, Weixin Luo, Dongze Lian, and Shenghua Gao.
\newblock Future frame prediction for anomaly detection - {A} new baseline.
\newblock In {\em {CVPR}}, pages 6536--6545. Computer Vision Foundation /
  {IEEE} Computer Society, 2018.

\bibitem{DBLP:conf/cvpr/LiuHBDBT21}
Xiaolong Liu, Yao Hu, Song Bai, Fei Ding, Xiang Bai, and Philip H.~S. Torr.
\newblock Multi-shot temporal event localization: {A} benchmark.
\newblock In {\em {CVPR}}, pages 12596--12606. Computer Vision Foundation /
  {IEEE}, 2021.

\bibitem{DBLP:conf/aaai/LiuLWWTWLHL20}
Zhaoyang Liu, Donghao Luo, Yabiao Wang, Limin Wang, Ying Tai, Chengjie Wang,
  Jilin Li, Feiyue Huang, and Tong Lu.
\newblock Teinet: Towards an efficient architecture for video recognition.
\newblock In {\em {AAAI}}, pages 11669--11676. {AAAI} Press, 2020.

\bibitem{Liu_2021_ICCV}
Zhaoyang Liu, Limin Wang, Wayne Wu, Chen Qian, and Tong Lu.
\newblock Tam: Temporal adaptive module for video recognition.
\newblock In {\em Proceedings of the IEEE/CVF International Conference on
  Computer Vision (ICCV)}, pages 13708--13718, October 2021.

\bibitem{DBLP:conf/nips/LuBPL19}
Jiasen Lu, Dhruv Batra, Devi Parikh, and Stefan Lee.
\newblock Vilbert: Pretraining task-agnostic visiolinguistic representations
  for vision-and-language tasks.
\newblock In {\em NeurIPS}, pages 13--23, 2019.

\bibitem{DBLP:conf/eccv/LuYR020}
Yiwei Lu, Frank Yu, Mahesh Kumar~Krishna Reddy, and Yang Wang.
\newblock Few-shot scene-adaptive anomaly detection.
\newblock In {\em {ECCV} {(5)}}, volume 12350 of {\em Lecture Notes in Computer
  Science}, pages 125--141. Springer, 2020.

\bibitem{DBLP:conf/iccv/LuoY19}
Chenxu Luo and Alan~L. Yuille.
\newblock Grouped spatial-temporal aggregation for efficient action
  recognition.
\newblock In {\em {ICCV}}, pages 5511--5520. {IEEE}, 2019.

\bibitem{DBLP:conf/cvpr/QingSGW0W0YGS21}
Zhiwu Qing, Haisheng Su, Weihao Gan, Dongliang Wang, Wei Wu, Xiang Wang, Yu
  Qiao, Junjie Yan, Changxin Gao, and Nong Sang.
\newblock Temporal context aggregation network for temporal action proposal
  refinement.
\newblock In {\em {CVPR}}, pages 485--494. Computer Vision Foundation / {IEEE},
  2021.

\bibitem{DBLP:conf/cvpr/ShaoZDL20}
Dian Shao, Yue Zhao, Bo Dai, and Dahua Lin.
\newblock Intra- and inter-action understanding via temporal action parsing.
\newblock In {\em {CVPR}}, pages 727--736. Computer Vision Foundation / {IEEE},
  2020.

\bibitem{Shou_2021_ICCV}
Mike~Zheng Shou, Stan~Weixian Lei, Weiyao Wang, Deepti Ghadiyaram, and Matt
  Feiszli.
\newblock Generic event boundary detection: A benchmark for event segmentation.
\newblock In {\em Proceedings of the IEEE/CVF International Conference on
  Computer Vision (ICCV)}, pages 8075--8084, October 2021.

\bibitem{DBLP:conf/nips/SimonyanZ14}
Karen Simonyan and Andrew Zisserman.
\newblock Two-stream convolutional networks for action recognition in videos.
\newblock In {\em {NIPS}}, pages 568--576, 2014.

\bibitem{DBLP:conf/cvpr/SultaniCS18}
Waqas Sultani, Chen Chen, and Mubarak Shah.
\newblock Real-world anomaly detection in surveillance videos.
\newblock In {\em {CVPR}}, pages 6479--6488. Computer Vision Foundation /
  {IEEE} Computer Society, 2018.

\bibitem{sun2021generic}
Quan Sun, Shimin Chen, Chen Chen, Xunqiang Tao, and Yandong Guo.
\newblock Generic event boundary detection: Submission to loveu challenge 2021.
\newblock
  \url{https://github.com/VisualAnalysisOfHumans/LOVEU_TRACK1_TOP3_SUBMISSION},
  2021.

\bibitem{Tan_2021_ICCV}
Jing Tan, Jiaqi Tang, Limin Wang, and Gangshan Wu.
\newblock Relaxed transformer decoders for direct action proposal generation.
\newblock In {\em Proceedings of the IEEE/CVF International Conference on
  Computer Vision (ICCV)}, pages 13526--13535, October 2021.

\bibitem{DBLP:conf/accv/TangFKCZ18}
Shitao Tang, Litong Feng, Zhanghui Kuang, Yimin Chen, and Wei Zhang.
\newblock Fast video shot transition localization with deep structured models.
\newblock In {\em {ACCV} {(1)}}, volume 11361 of {\em Lecture Notes in Computer
  Science}, pages 577--592. Springer, 2018.

\bibitem{DBLP:conf/eccv/TianSLDX18}
Yapeng Tian, Jing Shi, Bochen Li, Zhiyao Duan, and Chenliang Xu.
\newblock Audio-visual event localization in unconstrained videos.
\newblock In {\em {ECCV} {(2)}}, volume 11206 of {\em Lecture Notes in Computer
  Science}, pages 252--268. Springer, 2018.

\bibitem{DBLP:conf/iccv/TranWFT19}
Du Tran, Heng Wang, Matt Feiszli, and Lorenzo Torresani.
\newblock Video classification with channel-separated convolutional networks.
\newblock In {\em {ICCV}}, pages 5551--5560. {IEEE}, 2019.

\bibitem{tversky2013event}
Barbara Tversky and Jeffrey~M Zacks.
\newblock Event perception.
\newblock {\em Oxford handbook of cognitive psychology}, 2013.

\bibitem{DBLP:conf/nips/VaswaniSPUJGKP17}
Ashish Vaswani, Noam Shazeer, Niki Parmar, Jakob Uszkoreit, Llion Jones,
  Aidan~N. Gomez, Lukasz Kaiser, and Illia Polosukhin.
\newblock Attention is all you need.
\newblock In {\em {NIPS}}, pages 5998--6008, 2017.

\bibitem{DBLP:conf/cvpr/0002TJW21}
Limin Wang, Zhan Tong, Bin Ji, and Gangshan Wu.
\newblock {TDN:} temporal difference networks for efficient action recognition.
\newblock In {\em {CVPR}}, pages 1895--1904. Computer Vision Foundation /
  {IEEE}, 2021.

\bibitem{DBLP:conf/eccv/WangXW0LTG16}
Limin Wang, Yuanjun Xiong, Zhe Wang, Yu Qiao, Dahua Lin, Xiaoou Tang, and
  Luc~Van Gool.
\newblock Temporal segment networks: Towards good practices for deep action
  recognition.
\newblock In {\em {ECCV} {(8)}}, volume 9912 of {\em Lecture Notes in Computer
  Science}, pages 20--36. Springer, 2016.

\bibitem{DBLP:conf/icml/XuBKCCSZB15}
Kelvin Xu, Jimmy Ba, Ryan Kiros, Kyunghyun Cho, Aaron~C. Courville, Ruslan
  Salakhutdinov, Richard~S. Zemel, and Yoshua Bengio.
\newblock Show, attend and tell: Neural image caption generation with visual
  attention.
\newblock In {\em {ICML}}, volume~37 of {\em {JMLR} Workshop and Conference
  Proceedings}, pages 2048--2057. JMLR.org, 2015.

\bibitem{DBLP:conf/cvpr/XuZRTG20}
Mengmeng Xu, Chen Zhao, David~S. Rojas, Ali~K. Thabet, and Bernard Ghanem.
\newblock {G-TAD:} sub-graph localization for temporal action detection.
\newblock In {\em {CVPR}}, pages 10153--10162. Computer Vision Foundation /
  {IEEE}, 2020.

\bibitem{DBLP:conf/cvpr/YangHGDS16}
Zichao Yang, Xiaodong He, Jianfeng Gao, Li Deng, and Alexander~J. Smola.
\newblock Stacked attention networks for image question answering.
\newblock In {\em {CVPR}}, pages 21--29. {IEEE} Computer Society, 2016.

\bibitem{DBLP:conf/iccv/ZengHGTRZH19}
Runhao Zeng, Wenbing Huang, Chuang Gan, Mingkui Tan, Yu Rong, Peilin Zhao, and
  Junzhou Huang.
\newblock Graph convolutional networks for temporal action localization.
\newblock In {\em {ICCV}}, pages 7093--7102. {IEEE}, 2019.

\bibitem{DBLP:journals/pami/ZhangT12}
Zhang Zhang and Dacheng Tao.
\newblock Slow feature analysis for human action recognition.
\newblock {\em {IEEE} Trans. Pattern Anal. Mach. Intell.}, 34(3):436--450,
  2012.

\bibitem{DBLP:conf/cvpr/ZhaoXL18}
Yue Zhao, Yuanjun Xiong, and Dahua Lin.
\newblock Recognize actions by disentangling components of dynamics.
\newblock In {\em {CVPR}}, pages 6566--6575. Computer Vision Foundation /
  {IEEE} Computer Society, 2018.

\bibitem{Zhi_2021_ICCV}
Yuan Zhi, Zhan Tong, Limin Wang, and Gangshan Wu.
\newblock Mgsampler: An explainable sampling strategy for video action
  recognition.
\newblock In {\em Proceedings of the IEEE/CVF International Conference on
  Computer Vision (ICCV)}, pages 1513--1522, October 2021.

\end{thebibliography}
}

\end{document}